\documentclass[letterpaper]{article} %

\usepackage[preprint]{aaai2027}   %

\usepackage[hyphens]{url}  %
\usepackage{graphicx} %
\usepackage{natbib}  %
\usepackage{caption} %
\usepackage{algorithm}
\usepackage{algorithmic}
\usepackage{soul}
\usepackage{newfloat}
\usepackage{listings}
\DeclareCaptionStyle{ruled}{labelfont=normalfont,labelsep=colon,strut=off} %
\floatstyle{ruled}
\newfloat{listing}{tb}{lst}{}
\floatname{listing}{Listing}

\usepackage{booktabs}

\definecolor{scoredark}{HTML}{8B0000}  %
\definecolor{scorebg}{HTML}{8B0000}
\definecolor{darkgreen}{rgb}{0,0.5,0}
\definecolor{darkred}{rgb}{0.7,0,0}
\definecolor{teal}{rgb}{0.1,0.6,0.7}
\definecolor{blue}{rgb}{0.0,0.1,0.9}
\definecolor{orange}{rgb}{1.,0.7,0.0}
\definecolor{palegreen}{rgb}{0.7,0.7,0.0}
\definecolor{lightblue}{rgb}{0.70, 0.80, 0.89}
\definecolor{violet}{rgb}{0.50, 0.16, 0.88}
\definecolor{babyblue}{rgb}{0.00, 0.88, 0.88}
\definecolor{electricpurple}{rgb}{0.75, 0.0, 1.0}

\definecolor{casebg}{RGB}{255,252,225}
\definecolor{caseframe}{RGB}{120,110,70}
\definecolor{casehighlight}{RGB}{255,231,140}
\definecolor{querybg}{RGB}{235,246,255}
\definecolor{queryframe}{RGB}{95,130,155}

\newcommand{\hlcase}[1]{%
  \begingroup
  \setlength{\fboxsep}{0.4pt}%
  \colorbox{casehighlight}{\bfseries #1}%
  \endgroup
}

\lstdefinestyle{casequery}{
    style=casestudy,
    backgroundcolor=\color{querybg},
    rulecolor=\color{queryframe},
    moredelim=**[is][\bfseries]{@@}{@@}
}

\lstdefinestyle{casestudy}{
    basicstyle=\ttfamily\footnotesize,
    backgroundcolor=\color{casebg},
    frame=single,
    rulecolor=\color{caseframe},
    numbers=none,
    breaklines=true,
    breakatwhitespace=true,
    breakautoindent=false,
    breakindent=0pt,
    columns=fullflexible,
    keepspaces=false,
    showspaces=false,
    showtabs=false,
    showstringspaces=false,
    tabsize=2,
    autogobble=true,
    framesep=5pt,
    xleftmargin=0pt,
    xrightmargin=0pt,
    aboveskip=5pt,
    belowskip=8pt,
    escapeinside={(*@}{@*)}
}

\definecolor{promptbg}{RGB}{248,248,248}
\definecolor{promptframe}{RGB}{140,140,140}
\lstdefinestyle{llmprompt}{
    basicstyle=\ttfamily\scriptsize,
    backgroundcolor=\color{promptbg},
    frame=single,
    rulecolor=\color{promptframe},
    numbers=none,
    breaklines=true,
    breakatwhitespace=true,
    breakautoindent=false,
    breakindent=0pt,
    columns=fullflexible,
    keepspaces=true,
    showspaces=false,
    showtabs=false,
    showstringspaces=false,
    tabsize=2,
    autogobble=true,
    framesep=4pt,
    xleftmargin=0pt,
    xrightmargin=0pt,
    aboveskip=6pt,
    belowskip=10pt
}

\newcount\Comments  %
\newcommand{\kibitz}[2]{\ifnum\Comments=1{{\textcolor{#1}{\textsf{\footnotesize [#2]}}}}\fi}

\DeclareRobustCommand{\method}{\texttt{Baikal}\xspace}

\definecolor{algcomment}{HTML}{666666}

\newcommand{\mci}[2]{#1{\tiny\,$\pm$\,#2}}
\newcommand{\mcib}[2]{\textbf{#1}{\tiny\,$\pm$\,#2}}
\newcommand{\mciu}[2]{\underline{#1}{\tiny\,$\pm$\,#2}}

\usepackage{xcolor}
\usepackage{colortbl}
\usepackage{xspace}
\usepackage{makecell}
\usepackage{amssymb}
\usepackage{lstautogobble}
\usepackage{multirow}
\usepackage{subfigure}
\usepackage{xr}
\ifpreprint\else
\fi

\title{
\ifpreprint
    \raisebox{-0.7ex}{\includegraphics[height=1.5em]{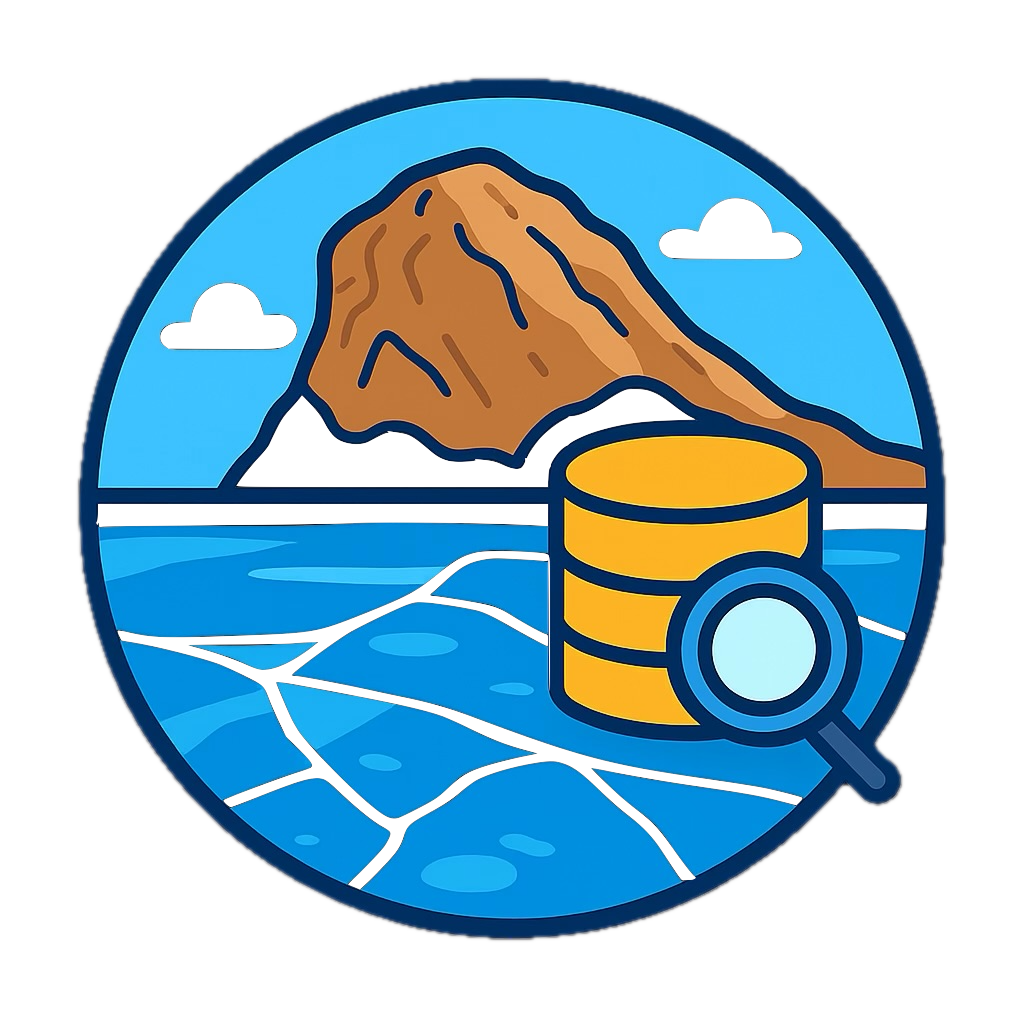}}\,
\fi
\method: Structured Search for Deep Research over Data Lakes
}
\author {
    Dhruv Agarwal\textsuperscript{\rm 1}\corresponding,
    Rishitha Guttapalle Mohan\textsuperscript{\rm 1},
    Aarti Kumari\textsuperscript{\rm 1},
    Ashi Sinha\textsuperscript{\rm 1},
    Athulya Anil\textsuperscript{\rm 1},
    Kavitha Srinivas\textsuperscript{\rm 2},
    Horst Samulowitz\textsuperscript{\rm 2},
    Andrew McCallum\textsuperscript{\rm 1}
}
\affiliations {
    \textsuperscript{\rm 1}University of Massachusetts Amherst\\
    \textsuperscript{\rm 2}IBM Research\\
    \{dagarwal, mccallum\}@cs.umass.edu, \{kavitha.srinivas, samulowitz\}@ibm.com\\
    \ifpreprint
        \vspace{1em}
        \raisebox{-0.15ex}{\includegraphics[height=0.9em]{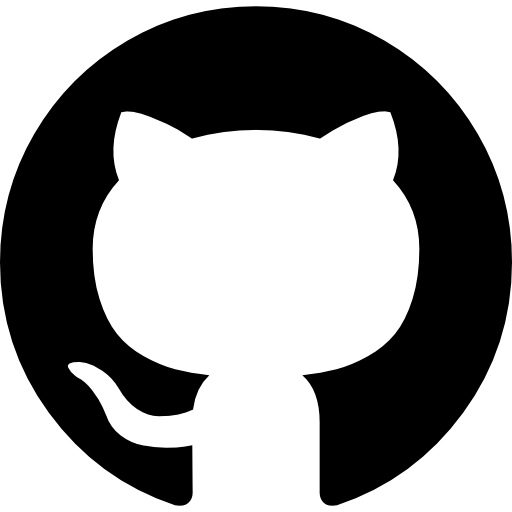}}\,\textbf{Code:} \url{https://github.com/dhdhagar/baikal}
    \fi
}

\begin{document}

\ifpreprint
  \makeatletter
  \let\baikal@label\label
  \renewcommand{\label}[1]{%
    \baikal@label{#1}%
    \protected@write\@auxout{}{%
      \string\newlabel{S-#1}{{\@currentlabel}{\thepage}}}%
    \protected@write\@auxout{}{%
      \string\newlabel{M-#1}{{\@currentlabel}{\thepage}}}%
  }%
  \makeatother
\fi

\maketitle

\begin{abstract}
Deep research over data lakes requires an LLM agent to selectively investigate evidence distributed across thousands of heterogeneous tables and passages to synthesize a report.
Existing methods perform iterative retrieval and generation, allowing the accumulating context to determine what to investigate next, which can overexploit locally promising evidence and fail to cover distinct semantic regions under a fixed budget.
To address this, we cast deep research over data lakes as a budgeted \emph{search} problem and present \method---a framework that first constructs a structured search space by clustering heterogeneous evidence into semantic regions, then 
performs adaptive, sequential search
to balance exploration and exploitation. 
Within each selected region, \method generates and investigates region-grounded subquestions, 
using the quality of the resulting findings as rewards to update region-level value estimates and guide subsequent search decisions
under policies ranging from random and LLM-guided selection to Bayesian $\epsilon$-greedy and UCB.
We evaluate \method on 15 deep research queries each over HybridQA and TAT-QA data lakes containing 10{,}993 and 2{,}757 tables, respectively, together with 227K Wikipedia passages and 13K financial-report passages.
To assess research quality, we introduce a rubric covering groundedness, relevance, distinctness, and utility, and use GPT-5-mini to score the outputs of \method and strong baselines, including DeepSearcher and an OpenCode research agent with retrieval and clustering variants.
Across both data lakes, \method performs strongly under several region-selection policies; its best-performing configuration improves report scores over the strongest baselines by 28\% on HybridQA and 36\% on TAT-QA.
Our analyses attribute these gains to the explicit organization and exploration of semantic evidence regions, which improves groundedness and distinctness and yields more useful findings under the same subquestion budget.
Together, these results demonstrate the value of structured semantic exploration for systematic research and discovery over heterogeneous data lakes.
\end{abstract}

\begin{figure*}[t]
    \centering
    \includegraphics[width=\textwidth]{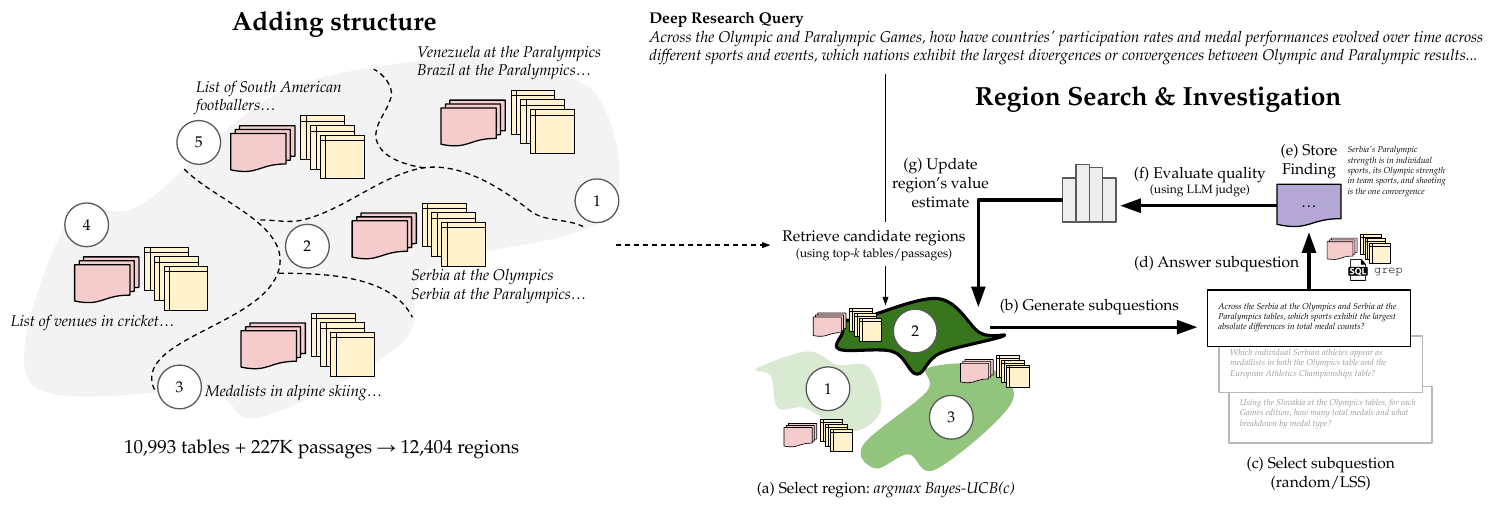}
    \caption{\textbf{Overview of \method}, on a real query from our HybridQA testbed.
    \emph{Structure} (once per lake): one encoder embeds every table and passage, and clustering groups them into semantic \emph{regions}---here 10{,}993 tables and 227K passages into 12{,}404 regions (\S\ref{sec:method-search-space}).
    \emph{Candidates} (per query): every region holding a top-$k$ retrieved item is retained, so one hit pulls in the rest of its region, placing \emph{Serbia at the Olympics} and \emph{Serbia at the Paralympics} together.
    \emph{Search} (per step): (a) a policy picks a region (here Bayes-UCB); (b,~c) an LLM proposes subquestions grounded in it and one is selected (random or LSS); (d,~e) a coding agent answers it with region-scoped SQL and passages, optionally widening the region via \texttt{grep}, then stores the finding; (f) a judge scores the finding on groundedness, relevance, distinctness, and utility; and (g) that score is used to update the policy's region estimate for subsequent selection. After $B$ steps the findings are synthesized into a long-form report using an LLM (\S\ref{sec:method-selection}--\ref{sec:method-report}).}
    \label{fig:method-overview}
\end{figure*}

\section{Introduction}
\label{sec:intro}
Large language models (LLMs) are increasingly being deployed in complex, long-horizon tasks that require autonomous decision-making, such as in software engineering~\citep{anthropic2025claudecode,openai2025codex,jiang2025aide} and scientific discovery~\citep{aiscientist,aicoscientist,mitchener2025kosmos}.
\emph{Deep research} is a canonical instance of such a task, where an agent must decompose an open-ended analytical query, search a large evidence collection with high fan-out over concepts, and synthesize what it gathers into a report whose claims are attributed to their sources~\citep{java2025livedrbench}.
Commercial systems~\citep{openaideep2025research,googledeep2025research} and their open counterparts~\citep{li2025webthinker,zilliz2025deepsearcher} typically target the open web~\citep{skarlinski2024language}, but for many organizations, the evidence sits in proprietary, internal storage, such as a \emph{data lake}---thousands of relational tables alongside the unstructured text that describes them~\citep{Brickley2019GoogleDS,Chakraborty2024NavigatorAG}.
\emph{Deep research over data lakes} is thus an important variant of the task, which recent benchmarks operationalize by turning table-text collections into open-world analytical requests~\citep{chowdhury2025dp,zhang2026dpdisc}.

Doing well on this task requires three capabilities: (a) \emph{retrieving} query-relevant evidence from heterogeneous data, (b) \emph{processing} the evidence with appropriate reasoning and analyses, and (c) \emph{finding} a diverse set of salient aspects for the query, often under a finite budget.
While there has been much progress on the first two---with text-to-SQL~\citep{Sun2023SQLPaLMIL, Agarwal2023BringYO} and coding agents~\citep{anthropic2025claudecode,openai2025codex}---the third aspect of \emph{search} requires further attention. 
Most existing systems~\citep{datadrivendiscovery,gu2024bladebenchmarkinglanguagemodel} rely on accumulating context to shift model beliefs~\citep{geng2025accumulating,agarwal2026evidence} in order to decide what to investigate next, which results in overexploiting locally-promising evidence~\citep{krishnamurthy2024can}\footnote{In deep research, this shows up as low-quality reports with poor coverage of a query's salient aspects (\S\ref{sec:results-main}).}. 

We therefore recast deep research over data lakes as a \emph{budgeted search} problem and present \textbf{\method}---a method that first structures the lake by jointly embedding~\citep{zhang2025qwen3embedding} and clustering~\citep{grootendorst2022bertopic} thousands of tables and passages into semantic \emph{regions}, followed by searching over them as a budgeted bandit problem~\citep{rlbook}: select a region using a policy, generate a grounded subquestion using an LLM, and answer the subquestion using region-scoped SQL and passages.
An LLM judge~\citep{zheng2023judging} scores the resulting finding on groundedness, relevance, distinctness, and utility, serving as a search reward.
We study policies from random and LLM-guided selection to Bayes $\epsilon$-greedy and Bayes-UCB~\citep{kaufmann2012bayesucb} with LLM-elicited region priors~\citep{agarwal2025autodiscovery}, and evaluate on a new testbed of lakes derived from HybridQA~\citep{chen-etal-2020-hybridqa} and TAT-QA~\citep{zhu-etal-2021-tat}.
We find that \method improves report score over the strongest baselines by 28\% and 36\% on the two lakes, and handing the same regions to a strong coding agent does not transfer these gains---the advantage comes from searching over the structure, not from the structure alone.

In summary, our contributions are as follows:
\begin{itemize}
\itemsep 0pt
\item We formalize deep research over data lakes as a budgeted search problem (\S\ref{sec:method-problem}).
\item We present \method, which structures a lake into semantic regions and conducts search using finding quality (\S\ref{sec:method}).
\item We build a testbed of deep-research queries over HybridQA and TAT-QA, with a practical rubric to evaluate quality using LLM judges (\S\ref{sec:experiments}).
\item We show through ablations and baseline comparisons that gains come from search rather than structure alone (\S\ref{sec:results}).
\item We further report inter-LLM agreement, cost analyses, retrieval quality, effect of LLM priors and scaling, and all prompts in \S\ref{sec:results-analyses} and the supplementary (\S\ref{S-app:full-ablations}--\ref{S-app:llm-prompts}).
\end{itemize}

\section{Related Work}
\label{sec:related}

\paragraph{Deep research.}
Deep-research systems~\citep{openaideep2025research,googledeep2025research,li2025webthinker,zilliz2025deepsearcher,java2025livedrbench} answer analytical queries by interleaving retrieval, reasoning, and report writing, querying an unstructured corpus one step at a time.
Recent benchmarks port the task to table-text lakes~\citep{chowdhury2025dp,zhang2026dpdisc}, but contribute evaluation rather than a search method.
We instead impose structure over the lake and treat the choice of where to look next as an explicit budgeted decision.

\paragraph{LLM-based search.}
A growing line of work casts LLM generation as search over a solution space: OPRO~\citep{yang2024opro} proposes successively better solutions from a trajectory of scored candidates, FunSearch~\citep{funsearch} and AlphaEvolve~\citep{novikov2025alphaevolve} evolve programs against automated evaluators, tree search has been applied to web agents~\citep{treesearch}, and closer to our policy-driven formulation, BOPRO~\citep{agarwal2025searching}, MiGrATe~\citep{phan2025migrate}, and TTT-Discover~\citep{yuksekgonul2026learning} search at test time with Bayesian optimization or reinforcement learning.
These methods search over candidate \emph{solutions} to a well-specified objective; we search over \emph{regions of a data lake} to decide where evidence should be gathered, with the reward supplied by a finding-quality rubric rather than a task metric.

\paragraph{Autonomous research and discovery.}
Agentic systems now execute substantial parts of the research process, from ideation and experimentation~\citep{aiscientist,yamada2025aiscientistv2workshoplevelautomated,aicoscientist,jansen2025codescientist} to autonomous laboratory work~\citep{boiko2023emergentautonomousscientificresearch} and open-ended exploration of a \emph{single} dataset~\citep{datadrivendiscovery,agarwal2025autodiscovery}.
Our setting differs in the search space rather than the goal: the relevant evidence must first be located within a heterogeneous lake of thousands of tables and passages, then searched over under a fixed budget.

\section{Method}
\label{sec:method}

\subsection{Problem Formulation}
\label{sec:method-problem}

A data lake $\mathcal{L} = \mathcal{T} \cup \mathcal{P}$ holds relational tables $\mathcal{T}$ and text passages $\mathcal{P}$ spanning many largely unrelated topics.
Given a research query $q$, a deep-research system must produce a report grounded in $\mathcal{L}$ under a fixed budget of $B$ steps, where each step contributes at most one \emph{finding}: a subquestion, its answer, and the lake evidence cited in support.
Let $f_t$ be the finding at step $t$ and $\rho(f_t \mid q, f_{1:t-1}) \in [0,1]$ be a scalar quality score that rewards findings that are grounded in lake evidence relevant to $q$, distinct from earlier findings, and useful in a report.
The objective, then, is to maximize the average finding score across budget steps, which we call the \textbf{report score}:
\begin{equation}
\label{eq:objective}
\max \; \frac{1}{B}\sum_{t=1}^{B} \rho\big(f_t \mid q, f_{1:t-1}\big).
\end{equation}

Two properties make this difficult.
First, the budget is small relative to the lake (e.g., $B = 50$ subquestions over $10^4$ tables and $10^5$ passages), so the choice of \emph{where} to look dominates end-to-end quality.
Second, the distinctness term makes the reward history-dependent: an evidence source loses value as it is mined, the non-stationarity inherent to any system that judges novelty against accumulated evidence~\citep{agarwal2026evidence}.
Iterative retrieve-and-generate agents choose each action from accumulated context, which biases them toward locally promising evidence and offers no mechanism for spreading a scarce budget across complementary parts of the lake.
\method instead treats exploration as an explicit budgeted search decision over a structured space of semantic evidence regions, and solves it with sequential decision-making under uncertainty~\citep{rlbook}.

\subsection{Adding Structure for Search}
\label{sec:method-search-space}

\paragraph{Evidence embeddings.}
Each table is encoded as text from its title, column names, and a schema description; each passage from its title and text.
A single encoder (Qwen3-Embedding-0.6B~\citep{zhang2025qwen3embedding} in our experiments) embeds both item types into a shared space, so that structured and unstructured evidence about the same topic lands in the same neighborhood.

\paragraph{Semantic regions.}
We organize the lake into a set of searchable regions $\mathcal{R}$ by jointly clustering table and passage embeddings with BERTopic~\citep{grootendorst2022bertopic}, which applies UMAP~\citep{mcinnes2018umap} for dimensionality reduction followed by density-based HDBSCAN clustering~\citep{mcinnes2017hdbscan}.
Noise assignments and clusters below a minimum size are discarded, and oversized clusters are split with $k$-means so that every retained region fits in an agent's context while remaining topically coherent.
A region $c$ therefore consists of tables $\mathcal{T}_c$, passages $\mathcal{P}_c$, and a short description derived from its representative terms.
Regions are the arms of the search problem: clustering reduces the $|\mathcal{L}|$ items of the lake to $|\mathcal{R}| \ll |\mathcal{L}|$ semantically labeled units, enabling credit assignment under a budget of $B$ steps.
Since clustering is query-independent, it is computed once per lake and amortized over all queries.

\paragraph{Query-conditioned candidates.}
At query time, \method embeds $q$, retrieves the top-$k$ tables and top-$k$ passages by cosine similarity, and retains the regions overlapping this retrieval set,
\begin{equation}
\label{eq:candidates}
\mathcal{C}(q) = \big\{ c \;\big|\; \big(\mathcal{T}_c \cap \mathrm{TopK}_{\mathrm{tbl}}(q)\big) \cup \big(\mathcal{P}_c \cap \mathrm{TopK}_{\mathrm{pas}}(q)\big) \neq \emptyset \big\}.
\end{equation}
Search is thus confined to regions that are retrieval-relevant, while a single retrieved item brings its whole region into scope, so \method can investigate co-clustered evidence that falls below the retrieval cutoff.
What this candidate set can reach is bounded by both retrieval depth and the clustering filters above, since items left as noise or dropped with undersized clusters stay out of scope.
At our default $k$, roughly half the gold tables and a third of the gold passages are reachable on TAT-QA, so reachability caps report quality; raising $k$ recovers far more of it than a larger encoder does (see Table~\ref{S-tab:tatqa-retrieval-analysis} in supplementary), but raises cost.

\begin{algorithm}[t]
\footnotesize
\caption{Deep research using \textbf{\method}}
\label{alg:baikal}
\begin{algorithmic}[1]
\REQUIRE query $q$, regions $\mathcal{R}$, budget $B$, policy $\pi$
\STATE $\mathcal{C} \gets$ regions in $\mathcal{R}$ overlapping $\mathrm{TopK}(q)$ \COMMENT{Eq.~\ref{eq:candidates}}
\STATE elicit priors $(\alpha_c, \beta_c)\ \forall c \in \mathcal{C}$ \COMMENT{for Bayesian $\pi$ only}
\FOR{$t = 1$ \TO $B$}
  \REPEAT
    \STATE $c \gets \pi(\mathcal{C})$
    \STATE $S_c \gets$ subquestions of $c$ not yet asked
    \IF{$S_c = \emptyset$}
      \STATE $S_c \gets$ generate up to $K$ new subquestions for $c$
    \ENDIF
    \IF{$S_c = \emptyset$}
      \STATE $\mathcal{C} \gets \mathcal{C} \setminus \{c\}$ \COMMENT{region/cluster attrition}
    \ENDIF
  \UNTIL{$S_c \neq \emptyset$ or $\mathcal{C} = \emptyset$}
  \STATE pick a single $s_t \in S_c$; $S_c \gets S_c \setminus \{s_t\}$
  \STATE $f_t \gets$ investigate $s_t$ within $c$ via SQL and passages
  \STATE $r_t \gets \rho(f_t \mid q, f_{1:t-1})$ \COMMENT{LLM rubric}
  \STATE update $\pi$'s value estimate of $c$ with $r_t$ \COMMENT{Eq.~\ref{eq:posterior}}
\ENDFOR
\RETURN report synthesized from valid $f_{1:B}$
\end{algorithmic}
\end{algorithm}

\subsection{Investigating a Region}
\label{sec:method-investigate}

Each budget step investigates one selected region $c \in \mathcal{C}(q)$ and produces a single finding.

\paragraph{Subquestion generation and selection.}
An LLM proposes up to $K$ analytical subquestions answerable from $\mathcal{T}_c \cup \mathcal{P}_c$, prompted to cover distinct angles such as counts, comparisons, trends, and distributions, and conditioned on the subquestions already asked of that region.
The LLM filters candidates that duplicate previously answered subquestions, and the surviving pool persists with the region: revisiting a region consumes its remaining subquestions before new ones are generated.
An empty pool is the model's signal that the region is irrelevant to $q$, so the region is dropped from $\mathcal{C}(q)$ (\emph{region attrition}) and another region is selected within the same step, making an uninformative region cost a selection but not a finding.
\method then consumes exactly one subquestion $s_t$ from the pool per budget step, choosing it either uniformly at random or with an LLM policy that scores candidates by their expected finding quality (\emph{LLM subquestion selection}, LSS).

\paragraph{Subquestion execution.}
The subquestion is answered by a research agent restricted to the active region, wherein it may issue SQL against a region-scoped SQLite database, read region passages, and must return an answer citing the table and passage identifiers it used.
While this may be implemented using a SQL-generation loop with error correction, we find that using a coding agent such as OpenCode~\citep{AnomalyCo_OpenCode_2026} as the executor (\emph{research agent}, RA) has a better success rate, 
so we use this in our full configuration.
Additionally, structured regions may miss relevant passages due to query-level nuances that only become clear after a subquestion is generated. To address this, we allow 
growing the policy-selected region before answering a subquestion (\emph{region expansion}, RE). 
Concretely, given a subquestion, it proposes keywords, \texttt{grep} matches them against the lake's passage index, and a few newly matched passages are merged into $c$ for the current and subsequent steps.
This is the primitive that coding agents such as Claude Code~\citep{anthropic2025claudecode} and OpenAI Codex~\citep{openai2025codex} also use to navigate repositories too large to read in full, and it serves the same purpose over a lake\footnote{Ablations in Table~\ref{S-tab:baikal-ablations-full} (supplementary) show that ``$-$RA'' reduces report score by $0.05$ on average, and ``$-$RE'' by a further $0.03$.}.

\paragraph{Finding quality as reward.}
Each finding is scored by an LLM judge~\citep{zheng2023judging} against the rubric of Eq.~\ref{eq:objective}, yielding the reward $r_t = \rho(f_t \mid q, f_{1:t-1})$ used for search control.
Because the judge conditions on previously observed findings, rewards are non-stationary by construction: a region's estimated value falls once its distinctive content has been reported, which steers the policies below toward unexplored regions over time.
The policy observes only rewards for findings it has itself produced; no ground-truth supervision about which regions hold gold evidence enters the loop.
We validate this rubric against an independent LLM judge in \S\ref{sec:results}.

\subsection{Search via Region Selection}
\label{sec:method-selection}

Region selection is a budgeted multi-armed bandit over $\mathcal{C}(q)$ in which pulling arm $c$ means investigating one subquestion inside $c$ and observing $r_t$.
All policies share the same candidate set and the same investigation machinery, and differ only in how the next region is chosen.

\paragraph{LLM priors for the Bayesian policies.}
Uninformed, the Bayesian policies would waste the budget: each pull is one observation and $B$ is comparable to $|\mathcal{C}(q)|$,\footnote{In our experiments $B=50$ steps against $|\mathcal{C}(q)| \approx 43$ candidate regions on average, a count fixed by how many regions the top-$k$ retrieval touches and unrelated to the budget.} so a policy that begins in uniform ignorance can exhaust its entire budget sampling each region once, never accumulating enough evidence to tell regions apart.
\method therefore elicits $n$ categorical beliefs per candidate region from the LLM---how likely investigating $c$ is to yield a high-quality finding for $q$, following \citet{agarwal2025autodiscovery}---and maps them to fractional successes and failures over a uniform $\mathrm{Beta}(1,1)$, giving each region parameters $(\alpha_c, \beta_c)$ with mean $\mu_c = \alpha_c / (\alpha_c + \beta_c)$.
Observing reward $r_t$ from region $c$ updates its posterior with evidence weight $w$,
\begin{equation}
\label{eq:posterior}
\alpha_c \leftarrow \alpha_c + w\, r_t, \qquad
\beta_c \leftarrow \beta_c + w\, (1-r_t),
\end{equation}
a pseudo-count scheme that treats continuous rubric scores as soft Bernoulli evidence rather than an exact conjugate update.

\paragraph{Policies.}
We study four region selectors:
\begin{itemize}
\itemsep 0pt
\item \textbf{Random} samples $c$ uniformly from $\mathcal{C}(q)$. Despite ignoring rewards, it inherits the coverage induced by the region structure and is among the strongest policies in \S\ref{sec:results}.
\item \textbf{LLM policy} prompts the LLM with the research question, region descriptions, and each region's visit count and mean reward, and follows its choice, defaulting to random selection when a response cannot be parsed.\footnote{This fallback is rarely triggered: unlike random selection, the policy discards no regions to attrition (Table~\ref{S-tab:operational-main} in supplementary), so its behavior does not collapse to random.}
\item \textbf{Bayes $\epsilon$-greedy} exploits $\arg\max_c \mu_c$ with probability $1-\epsilon$, and otherwise explores by sampling from a softmax over posterior means, $\Pr(c) \propto \exp(\tau \mu_c)$, so that exploration remains prior-guided rather than uniform.
\item \textbf{Bayes-UCB}~\citep{kaufmann2012bayesucb} selects the region with the highest posterior upper quantile,
\begin{equation}
\label{eq:bayesucb}
c_t = \arg\max_{c \in \mathcal{C}(q)} \; F^{-1}_{\mathrm{Beta}(\alpha_c, \beta_c)}\!\left(1 - 1/t\right),
\end{equation}
where $F^{-1}$ is the Beta quantile function and $t$ is one plus the number of investigations so far, so the quantile tightens toward the posterior mean as the budget is consumed.
\end{itemize}
Without LLM priors, the Bayesian policies reduce to their classical counterparts driven by empirical means~\citep{auer2002finite}.
This reduction is degenerate here: the optimism term of classical UCB assigns infinite value to every unvisited arm, so when $B$ is comparable to $|\mathcal{C}(q)|$ the policy spends essentially its whole budget visiting regions once each in arbitrary order, which is random selection.
We therefore report classical UCB only through the prior ablation (Table~\ref{S-tab:llm-priors-ablation} in supplementary).

\subsection{Report Synthesis}
\label{sec:method-report}

Once the budget is exhausted---or $\mathcal{C}(q)$ empties through region attrition---\method synthesizes the collected findings into a concise report conditioned on $q$, instructing the LLM to attribute each claim to the supporting tables and analyses and to introduce no facts beyond the collected evidence.
Evaluation then scores both the individual findings and the budget-normalized aggregate of Eq.~\ref{eq:objective}.
Algorithm~\ref{alg:baikal} summarizes the full procedure.

\begin{table}[t]
    \centering
    \small
    \setlength{\tabcolsep}{4pt}
    \begin{tabular}{lcccc}
        \toprule
        \textbf{Lake} &
        \textbf{Tables} &
        \shortstack{\textbf{Raw}\\\textbf{Passages}} &
        \shortstack{\textbf{Synth.}\\\textbf{Passages}} &
        \textbf{Regions} \\
        \midrule
        HybridQA & 10,993 & 227,250 & 41,608 & 12,404 \\
        TAT-QA   &  2,757 &  13,155 &  4,760 &  1,299 \\
        \bottomrule
    \end{tabular}
    \caption{\textbf{Data lake statistics.} Number of relational tables, raw passages, synthetic passages, and semantic regions constructed jointly from tables and raw passages (\S\ref{sec:method-search-space}).}
    \label{tab:data-lake-statistics}
\end{table}

\section{Experiments}
\label{sec:experiments}

\subsection{Data and Metrics}
\label{sec:experiments-data}

\begin{table*}[t]
\centering
\footnotesize
\setlength{\tabcolsep}{3pt}
\renewcommand{\arraystretch}{1.15}
\begin{tabular}{@{}l
  cccc
  >{\columncolor{scorebg!12}}c
  cccc
  >{\columncolor{scorebg!12}}c
@{}}
\toprule
& \multicolumn{5}{c}{\textbf{HybridQA}} & \multicolumn{5}{c}{\textbf{TAT-QA}} \\
\cmidrule(lr){2-6} \cmidrule(lr){7-11}
\textbf{Method} & \textit{Grounded} & \textit{Relevance} & \textit{Distinct.} & \textit{Utility} & \textbf{Score} & \textit{Grounded} & \textit{Relevance} & \textit{Distinct.} & \textit{Utility} & \textbf{Score} \\
\midrule
OpenCode Agent
  & \mci{0.72}{0.13} & \mci{0.75}{0.10} & \mci{0.62}{0.10} & \mci{0.59}{0.07} & \mci{0.29}{0.08}
  & \mci{0.69}{0.16} & \mci{0.72}{0.15} & \mci{0.49}{0.14} & \mci{0.49}{0.12} & \mci{0.32}{0.09} \\
\quad + Retrieval
  & \mci{0.77}{0.12} & \mci{0.80}{0.09} & \mci{0.67}{0.09} & \mci{0.65}{0.07} & \mci{0.35}{0.06}
  & \mci{0.72}{0.14} & \mci{0.74}{0.14} & \mci{0.41}{0.11} & \mci{0.46}{0.09} & \mci{0.29}{0.09} \\
\quad + Clustering
  & \mci{0.73}{0.09} & \mciu{0.81}{0.09} & \mci{0.67}{0.10} & \mci{0.64}{0.07} & \mci{0.36}{0.07}
  & \mci{0.76}{0.15} & \mci{0.74}{0.14} & \mci{0.41}{0.08} & \mci{0.49}{0.10} & \mci{0.27}{0.06} \\
DeepSearcher
  & \mci{0.31}{0.07} & \mcib{0.97}{0.01} & \mci{0.66}{0.05} & \mci{0.29}{0.07} & \mci{0.27}{0.07}
  & \mci{0.38}{0.15} & \mcib{0.97}{0.01} & \mci{0.68}{0.08} & \mci{0.37}{0.14} & \mci{0.33}{0.13} \\

\addlinespace[1.5pt]
\arrayrulecolor{gray!80}
\midrule[0.01pt]
\arrayrulecolor{black}

\multicolumn{11}{@{}l@{}}{\textbf{\method} \tiny{(Ours)}} \\
\addlinespace[2pt]
\quad Random
  & \mcib{0.92}{0.02} & \mci{0.73}{0.03} & \mcib{0.86}{0.02} & \mci{0.68}{0.02} & \mciu{0.44}{0.04}
  & \mci{0.73}{0.04} & \mci{0.92}{0.02} & \mcib{0.70}{0.05} & \mci{0.65}{0.04} & \mcib{0.45}{0.04} \\
\quad LLM Policy
  & \mci{0.83}{0.03} & \mci{0.78}{0.03} & \mci{0.69}{0.05} & \mci{0.65}{0.02} & \mci{0.35}{0.04}
  & \mcib{0.78}{0.06} & \mciu{0.95}{0.01} & \mci{0.63}{0.03} & \mci{0.64}{0.04} & \mci{0.41}{0.04} \\
\quad Bayes $\epsilon$-greedy
  & \mci{0.90}{0.02} & \mci{0.77}{0.03} & \mci{0.81}{0.02} & \mciu{0.70}{0.02} & \mci{0.44}{0.03}
  & \mciu{0.76}{0.04} & \mci{0.92}{0.02} & \mci{0.66}{0.05} & \mcib{0.66}{0.04} & \mciu{0.45}{0.04} \\
\quad Bayes-UCB
  & \mciu{0.92}{0.02} & \mci{0.77}{0.03} & \mciu{0.85}{0.02} & \mcib{0.70}{0.02} & \mcib{0.46}{0.04}
  & \mci{0.72}{0.04} & \mci{0.93}{0.02} & \mciu{0.69}{0.04} & \mciu{0.65}{0.03} & \mci{0.45}{0.04} \\
\bottomrule
\end{tabular}
\caption{\textbf{Main results.} Report scores and rubric components on 15 deep-research queries per lake (GPT-5-mini backbone, budget 50).
\method rows use the full stack and differ only in the region-selection policy.
The shaded score is the budget-normalized sum of per-finding products (\textit{groundedness} $\times$ \textit{relevance} $\times$ \textit{distinctness} $\times$ \textit{utility}); other columns are query-level means of each component.
Best scores are \textbf{bold}, second best \underline{underlined}; $\pm$ is half the width of a 95\% bootstrap CI (10{,}000 paired query resamples).
Full protocol in \S\ref{sec:experiments}.}
\label{tab:main-results-raw}
\end{table*}

\paragraph{Data lakes.}
We build two lakes with different evidence characteristics (Table~\ref{tab:data-lake-statistics}).
\textbf{HybridQA}~\citep{chen-etal-2020-hybridqa} is an open-domain lake of Wikipedia tables and the articles linked from their cells; \textbf{TAT-QA}~\citep{zhu-etal-2021-tat} is a financial lake of report tables and their surrounding narrative paragraphs.
Tables are exposed as SQLite relations with generated schema descriptions, paired with the corresponding raw source passages.\footnote{DPDisc~\citep{zhang2026dpdisc} additionally supplies \emph{synthetic} passages in which an LLM summarizes each table row jointly with the articles it references; Table~\ref{S-tab:hybridqa-raw-vs-synth} in supplementary reports an evaluation under both passage forms. Unless noted, all experiments use raw passages.}

\paragraph{Building deep-research queries.}
We use DPDisc~\citep{zhang2026dpdisc}, a benchmark that repurposes table-text QA datasets into \emph{data product requests}: high-level analytical needs, each annotated with the collection of tables and passages required to satisfy it \footnote{We use these annotations as gold evidence for analysis only; no supervision about which regions hold gold evidence ever reaches the search loop.}.
From each lake we retain requests covering at least 20 gold tables, so that no query is answerable from a single region, and stratify the remainder by gold-table count into \textit{low} ($[20,25]$), \textit{medium} ($(25,35]$), and \textit{high} ($(35,72]$) coverage---a proxy for difficulty, since more gold tables means the information need is spread more widely across the lake.
An LLM rewrites each request as an analytical research question that states the information need without referring to any schema or table.
We evaluate on 15 queries per lake, balanced across the three difficulty strata.

\paragraph{Evaluation.}
An LLM judge scores every finding on the four rubric dimensions of Eq.~\ref{eq:objective}: \emph{groundedness} (binary), and \emph{relevance}, \emph{distinctness}, and \emph{utility} on a five-level ordinal scale mapped to $\{0, 0.25, 0.5, 0.75, 1\}$.
The judge sees the research question, the subquestion and its answer, the cited tables, SQL results, and passage text, together with all previously accepted findings, which is what makes distinctness history-dependent.
Answers that only report that information could not be found receive zero groundedness and utility.
A finding's score is the product of the four dimensions, so a finding must satisfy all of them to contribute, and the report score is the sum of finding scores divided by the budget, with steps that yield no finding contributing zero.
All confidence intervals are 95\% bootstrap intervals over $10{,}000$ paired query resamples.

\subsection{Baselines and Implementation}
\label{sec:experiments-baselines}

\paragraph{Baselines.}
Our primary baseline is \textbf{OpenCode Agent}~\citep{AnomalyCo_OpenCode_2026}, a canonical open-source coding-agent harness analogous to proprietary systems such as Claude Code~\citep{anthropic2025claudecode} and OpenAI Codex~\citep{openai2025codex}, given the whole lake: schema descriptions, a SQLite database over all tables, passage access, and shell search tools.
Two variants narrow its starting point using the same machinery \method uses: \textbf{+Retrieval} supplies the top-$k$ tables and passages for the query, and \textbf{+Clustering} additionally supplies the regions overlapping that retrieval set.
Comparing against these isolates the contribution of sequential region selection from that of retrieval and clustering alone.
We also compare against \textbf{DeepSearcher}~\citep{zilliz2025deepsearcher}, an iterative retrieve-and-reason agent over a vector index of the same lake rendered as text, which retrieves passages at every step and has no SQL access.
Every method receives the same budget of 50 steps and the same LLM backbone. 

\paragraph{Implementation.}
All methods use GPT-5-mini as the backbone and as the judge, and Qwen3-Embedding-0.6B for retrieval and clustering.
Regions are built with UMAP (10 neighbors, 10 components) and HDBSCAN (minimum size 3), retaining regions of 2--30 tables; this yields the 12{,}404 and 1{,}299 regions precomputed once per lake in Table~\ref{tab:data-lake-statistics}.
At query time we retrieve $k=100$ tables and $k=100$ passages, generate up to $K=8$ subquestions per region refill, and allow at most 3 SQL attempts per subquestion.
Bayesian policies draw $n=5$ prior samples per region, update posteriors with evidence weight $w=5$, and use $\epsilon = 0.1$ and softmax temperature $\tau = 1$.
The budget is $B=50$ findings, and each configuration is run once per query with a fixed seed of 42, so every reported number aggregates 15 runs and its confidence interval reflects variation across queries rather than across repeated runs.
Reported cost is total API spend per query from provider token accounting, covering every LLM and embedding call in a run, including the coding agent's own sessions and rubric scoring.

\section{Results and Analyses}
\label{sec:results}

\begin{figure*}[t]
    \centering
    \includegraphics[width=0.74\textwidth]{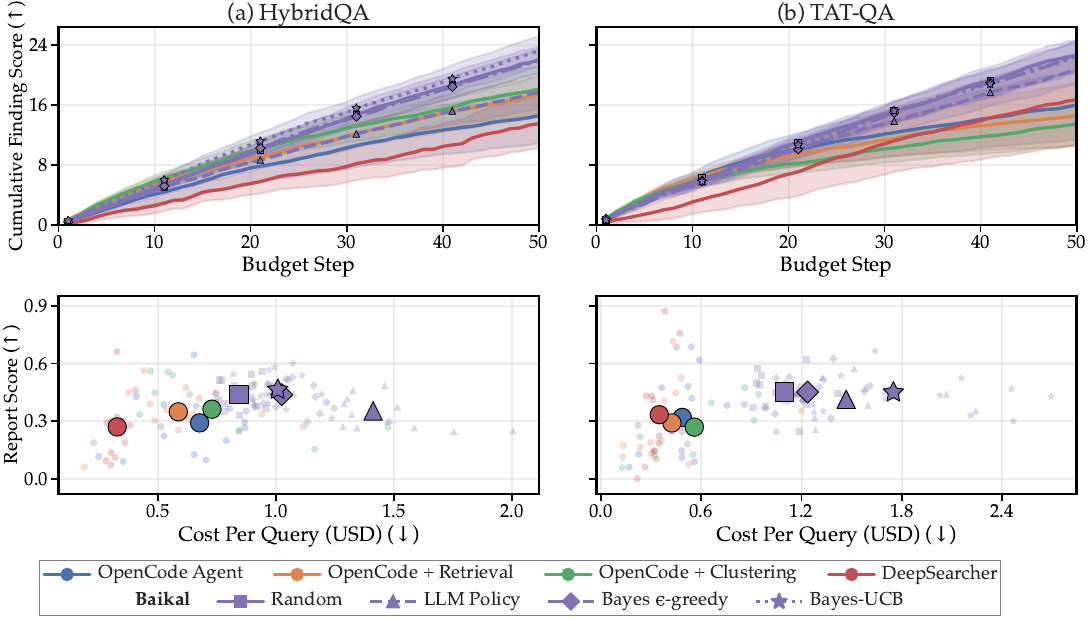}
    \caption{\textbf{Search trajectories and cost--quality trade-off.}
    \textit{Top:} cumulative finding score over the research budget; lines are mean trajectories across queries and shaded regions are 95\% confidence intervals.
    \textit{Bottom:} cost--quality trade-off, where small markers are individual queries and large outlined markers are method averages.
    Columns share a lake (HybridQA, TAT-QA) and all panels share the legend.}
    \label{fig:trajectory-cost-raw}
\end{figure*}

\subsection{Main Results}
\label{sec:results-main}

\textbf{\method outperforms baselines on both lakes.}
The best \method configuration reaches a report score of $0.46$ on HybridQA (Bayes-UCB) and $0.45$ on TAT-QA (random and Bayes $\epsilon$-greedy), improving on the strongest baseline for each lake by 28\% and 36\% respectively (Table~\ref{tab:main-results-raw}).
Three of the four policies beat every baseline on both lakes; only the LLM policy falls short on HybridQA ($0.35$), because it concentrates on the few regions it judges safe and forfeits coverage (Table~\ref{S-tab:operational-main} in supplementary).
\method's confidence intervals are also two to three times narrower, so the advantage holds across queries rather than resting on a few easy ones.

\textbf{Access to regions does not transfer gains to OpenCode.}
Giving the OpenCode agent the same retrieval and clustering artifacts helps on HybridQA ($0.29 \rightarrow 0.36$) but hurts on TAT-QA ($0.32 \rightarrow 0.27$), so handing a capable agent the structure is not by itself what separates \method from the baselines.

\textbf{Gains concentrate in groundedness and distinctness.}
On HybridQA, the strongest \method policies raise groundedness from $0.72$--$0.73$ for the OpenCode variants and $0.31$ for DeepSearcher to $0.92$, and distinctness from $0.62$--$0.67$ to $0.86$.
Relevance moves the other way: DeepSearcher scores highest on both lakes ($0.97$), and OpenCode + Clustering beats \method on HybridQA ($0.81$ vs.\ $0.73$).
This pattern is characteristic of iterative retrieve-and-generate search, which stays close to the query and reads as on-topic but is frequently unsupported by lake evidence; because the report score is multiplicative, relevance without groundedness does not improve it.

\subsection{Analyses}
\label{sec:results-analyses}

We now ask where the advantage comes from, what it costs, and whether the rubric that measures it is reproducible.

\textbf{The search procedure, not only the structure, produce the gains.}
Base \method---the same regions under a budgeted region-investigation loop with random selection, and without the research agent, region expansion, or LLM subquestion selection---already scores $0.36$ and $0.35$ (Table~\ref{S-tab:baikal-ablations-full} in supplementary), matching or beating the best baseline on both lakes.
Since OpenCode receives the same artifacts and does not reach this level, what separates the two is the sequential allocation of the budget over regions rather than the artifacts themselves.
This also explains why random selection is already among the strongest policies in Table~\ref{tab:main-results-raw}: at $B=50$ the structure and the loop do most of the work.

\textbf{\method recovers more of the gold evidence.}
If better budget allocation is the mechanism, \method should end the run having touched more of the evidence each query needs, and it does (Table~\ref{S-tab:retrieval-main} in supplementary).
It queries $0.32$--$0.35$ of gold tables on HybridQA against $0.14$--$0.18$ for the OpenCode variants and $0.03$ for DeepSearcher, and on TAT-QA cites $0.15$--$0.20$ of gold passages against $0.04$--$0.13$.
Passage recall on HybridQA is near zero for every method, which reflects scale rather than method: an average query has 749 gold passages and only 50 steps in which to reach them.

\begin{center}
    \includegraphics[width=0.8\columnwidth]{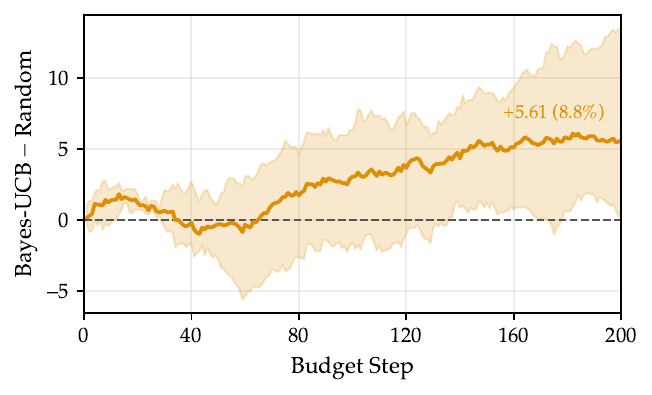}
    \vspace{-0.5em}
    \captionof{figure}{\textbf{Bayes-UCB vs. random search.}
    Paired difference in cumulative finding score between Bayes-UCB and Random through budget 200, averaged over three HybridQA queries spanning low, medium, and high coverage. Bayes-UCB achieves a 5.61-point (8.8\%) average gain at budget 200. Shading shows the 95\% bootstrap confidence interval across queries.}
    \label{fig:budget-trajectory-delta}
\end{center}

\textbf{\method's search advantage appears early, and its policies separate at larger budgets.}
Figure~\ref{fig:trajectory-cost-raw} (top) tracks cumulative finding score across the budget.
\method separates from every baseline within the first ten steps and the gap widens thereafter, so the advantage comes from where each step is spent rather than from a final synthesis pass.
No method flattens by step 50, so the lakes are far from exhausted at this budget.
The \method policies, however, are hard to tell apart at $B=50$, and this is a property of the budget rather than evidence that the policy does not matter: with $B$ comparable to $|\mathcal{C}(q)|$, a policy barely touches each region once and has little left to exploit.
Removing the LLM priors that make the first visit informative drops Bayes $\epsilon$-greedy to baseline level (Table~\ref{S-tab:llm-priors-ablation} in supplementary), which is consistent with this reading.
Extending the budget to 200 on three HybridQA queries spanning the coverage strata does separate them: Bayes-UCB gains $5.61$ points ($8.8\%$) in cumulative finding score over random (Figure~\ref{fig:budget-trajectory-delta}), so the comparison at $B=50$ understates rather than flatters the value of optimistic region selection.

\textbf{\method spends more per query because it produces more valid findings.}
Out of 50 steps on HybridQA, \method yields $44.6$ findings with a positive rubric score, against $30.9$ for OpenCode and $15.2$ for DeepSearcher (Table~\ref{S-tab:operational-main} in supplementary), and the baselines are far more variable across queries ($\pm 3.8$--$7.4$ versus $\pm 1.3$--$2.8$).
That higher yield shows up as higher total spend: \$0.84--\$1.41 per query for \method versus \$0.33--\$0.73 for the baselines, with a larger premium on TAT-QA (Figure~\ref{fig:trajectory-cost-raw}, bottom).
Normalized by valid findings, however, the costs align (\$0.019 for random \method vs.\ \$0.017--\$0.022 for the baselines), so the premium buys more usable evidence rather than paying for a more expensive step.
Within \method, random selection reaches $0.44$ at \$0.84 while Bayes-UCB reaches $0.46$ at roughly 20\% higher cost, and the LLM policy is dominated on both axes; at $B=50$, then, the cheapest policy is also close to the best (Figure~\ref{S-fig:model-scaling-cost-quality} in supplementary reports the same trade-off across LLMs).

\begin{center}
\setlength{\tabcolsep}{2.5pt}
\begin{tabular}{@{}lcccc@{}}
\toprule
\textbf{Dimension} & \textbf{Stat.} &
\textbf{Agree.} &
\textbf{Exact} & \textbf{Within-1} \\
\midrule
Groundedness & $\mathrm{AC}_1$ & $0.59\ [0.40, 0.75]$ & 76.2\% & -- \\
Relevance & $\mathrm{AC}_2$ & $0.82\ [0.74, 0.89]$ & 41.2\% & 85.0\% \\
Distinctness & $\mathrm{AC}_2$ & $0.91\ [0.86, 0.95]$ & 58.8\% & 92.5\% \\
Utility & $\mathrm{AC}_2$ & $0.33\ [0.13, 0.53]$ & 18.8\% & 61.3\% \\
\bottomrule
\end{tabular}
\captionof{table}{\textbf{Inter-LLM agreement on rubric.}
GPT-5-mini run-time judge vs.\ independent Claude Opus~4.6 on 80 score-stratified findings.
Gwet's $\mathrm{AC}_1$~\citep{gwet2008ac1} for groundedness; quadratic-weighted $\mathrm{AC}_2$ for ordinal dimensions (preferred under skewed prevalences); brackets are 95\% query-level bootstrap CIs.
\emph{Exact} / \emph{Within-1}: identical or adjacent categories.}
\label{tab:llm-rubric-validation}
\end{center}

\textbf{The finding-quality rubric reproduces across judges.}
Since every result above is scored by a GPT-5-mini judge, we re-scored 80 findings with an independent Claude Opus~4.6 judge on the same inputs (Table~\ref{tab:llm-rubric-validation}).
Agreement is broad but varies by dimension: it is highest on distinctness ($0.91$) and lowest on utility ($0.33$), the most subjective of the four though still well above chance on a $[-1,1]$ scale.
We report Gwet's $\mathrm{AC}$ coefficients because category prevalences here are skewed, under which Cohen's $\kappa$ can understate agreement; Table~\ref{S-tab:llm-rubric-validation-kappa} in supplementary reports $\kappa$/$\kappa_w$ for transparency.

\section{Conclusion}
\label{sec:conclusion}
We cast deep research over data lakes as a budgeted search problem and present \method, which clusters heterogeneous tables and passages into semantic regions and sequentially selects which region to investigate, scoring each finding to guide later decisions.
On a new testbed with automatic rubric scoring, \method improves report score over the strongest baseline on both lakes.
A strong coding agent given the same regions does not realize these gains, so the improvement stems from the search procedure rather than the structure alone: structure supplies coverage at small budgets, while the policy adds efficiency as the budget grows.
Treating a large, heterogeneous data lake as a structured space to be searched under an explicit budget, rather than a corpus to be retrieved from, offers a productive foundation for systematic research and discovery.

\bibliography{main}

\ifpreprint
    \clearpage
    \onecolumn
    \appendix
    \makeatletter
    \def\addcontentsline#1#2#3{%
      \addtocontents{#1}{\protect\contentsline{#2}{#3}{\thepage}{}%
                         \protected@file@percent}}
    \let\origlabel\label
    \def\label#1{\origlabel{#1}\origlabel{S-#1}}
    \makeatother
    \setcounter{tocdepth}{2}
    \renewcommand{\contentsname}{Appendix}
    \tableofcontents
    \clearpage
    \section{Ablations}
\label{app:full-ablations}

\begin{center}
\footnotesize
\setlength{\tabcolsep}{3.5pt}
\renewcommand{\arraystretch}{1.12}
\begin{tabular}{@{}l
  cccc
  >{\columncolor{scorebg!12}}c
  cccc
  >{\columncolor{scorebg!12}}c
@{}}
\toprule
& \multicolumn{5}{c}{\textbf{HybridQA}} & \multicolumn{5}{c}{\textbf{TAT-QA}} \\
\cmidrule(lr){2-6} \cmidrule(lr){7-11}
\textbf{Configuration} & \textit{Grounded} & \textit{Relevance} & \textit{Distinct.} & \textit{Utility} & \textbf{Score} & \textit{Grounded} & \textit{Relevance} & \textit{Distinct.} & \textit{Utility} & \textbf{Score} \\
\midrule
\textbf{\method} (w/ random) &
  \mci{0.92}{0.02} & \mci{0.73}{0.03} & \mci{0.86}{0.02} & \mci{0.68}{0.02} & \mci{0.44}{0.04} &
  \mci{0.73}{0.04} & \mci{0.92}{0.02} & \mci{0.70}{0.05} & \mci{0.65}{0.04} & \mci{0.45}{0.04} \\
\quad $-$ RA &
  \mci{0.83}{0.03} & \mci{0.63}{0.04} & \mci{0.76}{0.02} & \mci{0.60}{0.03} & \mci{0.40}{0.04} &
  \mci{0.57}{0.05} & \mci{0.74}{0.03} & \mci{0.56}{0.04} & \mci{0.48}{0.05} & \mci{0.39}{0.05} \\
\quad $-$ RA $-$ RE &
  \mci{0.80}{0.02} & \mci{0.58}{0.03} & \mci{0.72}{0.03} & \mci{0.56}{0.02} & \mci{0.37}{0.03} &
  \mci{0.53}{0.05} & \mci{0.70}{0.04} & \mci{0.55}{0.04} & \mci{0.45}{0.04} & \mci{0.36}{0.04} \\
\quad $-$ RA $-$ RE $-$ LSS &
  \mci{0.82}{0.03} & \mci{0.57}{0.04} & \mci{0.75}{0.04} & \mci{0.55}{0.03} & \mci{0.36}{0.04} &
  \mci{0.54}{0.06} & \mci{0.73}{0.05} & \mci{0.54}{0.05} & \mci{0.44}{0.05} & \mci{0.35}{0.04} \\
\bottomrule
\end{tabular}
\captionof{table}{\textbf{Full rubric breakdown of \method ablations.} The top row is the complete stack with LLM subquestion selection (LSS), region expansion (RE, dynamic passage grep into the active region), and a research agent (RA, OpenCode subquestion execution), using random region selection; it corresponds to the \emph{random} search row of Table~\ref{M-tab:main-results-raw} in the main paper. Each subsequent row removes the listed components cumulatively, and the bottom row is base \method (semantic regions with random subquestion selection). $\pm$ values give half the width of 95\% bootstrap CIs (10{,}000 paired resamples).}
\label{tab:baikal-ablations-full}
\end{center}

\textbf{Regions account for most of the score, and passage preprocessing helps weaker systems the most.}
Removing the research agent costs $0.04$ and $0.06$ report score on HybridQA and TAT-QA, region expansion a further $0.03$ on both, and LLM subquestion selection a final $0.01$ (Table~\ref{tab:baikal-ablations-full}); what remains---regions with random selection---still scores $0.36$ and $0.35$, so the rest of the stack adds $0.08$--$0.10$ on top of a foundation already as strong as the best baseline.

\begin{center}
\footnotesize
\setlength{\tabcolsep}{8pt}
\renewcommand{\arraystretch}{1.1}
\begin{tabular}{@{}l cc @{}}
\toprule
\textbf{Method} & \textbf{w/ Raw Passages} & \textbf{w/ Synth. Passages} \\
\midrule
\textbf{\method} (base) &
  \mci{0.36}{0.04} & \mci{0.38}{0.04} \\
OpenCode Agent &
  \mci{0.29}{0.08} & \mci{0.35}{0.07} \\
DeepSearcher &
  \mci{0.27}{0.07} & \mci{0.14}{0.06} \\
\bottomrule
\end{tabular}
\captionof{table}{\textbf{Effect of LLM preprocessing of lake passages.} Report scores for HybridQA deep-research queries (15 queries, GPT-5-mini, budget of 50 subquestion findings) when using \textit{raw} passages (full Wikipedia text) vs. \textit{synthetic} passages (LLM summaries of Wikipedia table rows jointly with raw texts of articles mentioned in those rows).}
\label{tab:hybridqa-raw-vs-synth}
\end{center}

Passage preprocessing shifts these numbers unevenly across systems: replacing raw source text with synthetic passages improves base \method modestly ($0.36 \rightarrow 0.38$) and OpenCode substantially ($0.29 \rightarrow 0.35$), while DeepSearcher degrades sharply ($0.27 \rightarrow 0.14$; Table~\ref{tab:hybridqa-raw-vs-synth}).
Summarization compresses the lake into fewer, denser passages, which helps agents that must read text to locate evidence but removes the redundancy that chunk-level vector retrieval depends on; that \method gains least suggests region structure already supplies most of the organization summarization provides.

\section{Additional Analyses}
\label{app:additional-analyses}

\subsection{Ground-Truth Evidence Usage}

\begin{center}
\footnotesize
\setlength{\tabcolsep}{3.5pt}
\renewcommand{\arraystretch}{1.15}
\begin{tabular}{@{}l
  cccc
  cccc
@{}}
\toprule
& \multicolumn{4}{c}{\textbf{HybridQA}} & \multicolumn{4}{c}{\textbf{TAT-QA}} \\
\cmidrule(lr){2-5} \cmidrule(lr){6-9}
& \multicolumn{2}{c}{\textbf{Tables} {\tiny [avg. 31.1/query]}}
& \multicolumn{2}{c}{\textbf{Passages} {\tiny [avg. 749.0/query]}}
& \multicolumn{2}{c}{\textbf{Tables} {\tiny [avg. 10.5/query]}}
& \multicolumn{2}{c}{\textbf{Passages} {\tiny [avg. 46.1/query]}} \\
\cmidrule(lr){2-3} \cmidrule(lr){4-5} \cmidrule(lr){6-7} \cmidrule(lr){8-9}
\textbf{Method}
  & \textbf{Recall} & \textbf{Prec.} & \textbf{Recall} & \textbf{Prec.}
  & \textbf{Recall} & \textbf{Prec.} & \textbf{Recall} & \textbf{Prec.} \\
\midrule
OpenCode Agent
  & \mci{0.14}{0.06} & \mci{0.17}{0.07}
  & \mci{0.00}{0.01} & \mci{0.72}{0.00}
  & \mci{0.21}{0.13} & \mci{0.12}{0.10}
  & \mci{0.04}{0.03} & \mci{0.49}{0.33} \\
\quad + Retrieval
  & \mci{0.18}{0.10} & \mci{0.19}{0.08}
  & \mci{0.01}{0.01} & \mci{0.65}{0.00}
  & \mci{0.16}{0.10} & \mci{0.29}{0.18}
  & \mci{0.13}{0.08} & \mci{0.53}{0.20} \\
\quad + Clustering
  & \mci{0.17}{0.07} & \mci{0.21}{0.07}
  & \mci{0.01}{0.01} & \mci{0.64}{0.00}
  & \mci{0.28}{0.14} & \mci{0.24}{0.09}
  & \mci{0.12}{0.08} & \mci{0.37}{0.09} \\
DeepSearcher
  & \mci{0.03}{0.04} & \mci{0.21}{0.20}
  & \mci{0.01}{0.00} & \mci{0.32}{0.00}
  & \mci{0.20}{0.08} & \mci{0.20}{0.12}
  & \mci{0.06}{0.03} & \mci{0.19}{0.29} \\
\arrayrulecolor{gray!80}
\midrule[0.01pt]
\arrayrulecolor{black}
\multicolumn{9}{@{}l@{}}{\textbf{\method} \tiny{(Ours)}} \\
\addlinespace[2pt]
\quad Random
  & \mci{0.33}{0.12} & \mci{0.14}{0.05}
  & \mci{0.02}{0.01} & \mci{0.34}{0.00}
  & \mci{0.25}{0.08} & \mci{0.08}{0.04}
  & \mci{0.15}{0.05} & \mci{0.12}{0.07} \\
\quad LLM Policy
  & \mci{0.18}{0.06} & \mci{0.17}{0.07}
  & \mci{0.00}{0.00} & \mci{0.62}{0.00}
  & \mci{0.40}{0.13} & \mci{0.23}{0.09}
  & \mci{0.07}{0.04} & \mci{0.22}{0.11} \\
\quad Bayes $\epsilon$-greedy
  & \mci{0.35}{0.13} & \mci{0.15}{0.05}
  & \mci{0.01}{0.01} & \mci{0.34}{0.00}
  & \mci{0.34}{0.13} & \mci{0.12}{0.08}
  & \mci{0.20}{0.07} & \mci{0.23}{0.08} \\
\quad Bayes-UCB
  & \mci{0.32}{0.11} & \mci{0.15}{0.05}
  & \mci{0.02}{0.01} & \mci{0.44}{0.00}
  & \mci{0.31}{0.11} & \mci{0.10}{0.06}
  & \mci{0.17}{0.05} & \mci{0.20}{0.12} \\
\bottomrule
\end{tabular}
\captionof{table}{\textbf{Ground-truth tables and passages used.} End-of-budget recall and precision of ground-truth tables among those queried via SQL, and of ground-truth passages among those cited.}
\label{tab:retrieval-main}
\end{center}

Table~\ref{tab:retrieval-main} reports end-of-budget recall and precision of annotated gold tables (among those queried via SQL) and gold passages (among those cited).
As discussed in the main paper, \method recovers substantially more gold tables on HybridQA and more gold passages on TAT-QA than the baselines; precision is lower because it touches more tables within the same budget, and HybridQA passage recall is near zero for every method given the scale of the gold set relative to the budget.

\subsection{Operational Metrics and Region Attrition}

\begin{center}
\footnotesize
\setlength{\tabcolsep}{5pt}
\renewcommand{\arraystretch}{1.15}
\begin{tabular}{@{}l
  ccc
  ccc
@{}}
\toprule
& \multicolumn{3}{c}{\textbf{HybridQA}} & \multicolumn{3}{c}{\textbf{TAT-QA}} \\
\cmidrule(lr){2-4} \cmidrule(lr){5-7}
\textbf{Method}
  & \textbf{SQL succ.} & \textbf{Attrition} & \textbf{\# Valid Findings}
  & \textbf{SQL succ.} & \textbf{Attrition} & \textbf{\# Valid Findings} \\
\midrule
OpenCode Agent
  & 0.99 & --- & \mci{30.9}{6.7}
  & 0.96 & --- & \mci{27.4}{7.4} \\
\quad + Retrieval
  & 0.99 & --- & \mci{35.4}{5.2}
  & 0.96 & --- & \mci{26.9}{6.7} \\
\quad + Clustering
  & 0.98 & --- & \mci{32.5}{4.9}
  & 0.97 & --- & \mci{31.7}{6.5} \\
DeepSearcher
  & --- & --- & \mci{15.2}{3.8}
  & --- & --- & \mci{18.5}{7.4} \\
\midrule
\multicolumn{7}{@{}l@{}}{\textbf{\method} \tiny{(Ours)}} \\
\addlinespace[2pt]
\quad Random
  & 0.87 & 0.23 & \mci{44.6}{1.3}
  & 0.83 & 0.12 & \mci{33.8}{1.9} \\
\quad LLM Policy
  & 0.97 & 0.00 & \mci{39.6}{1.7}
  & 0.96 & 0.00 & \mci{35.9}{2.8} \\
\quad Bayes $\epsilon$-greedy
  & 0.92 & 0.03 & \mci{43.1}{1.4}
  & 0.84 & 0.02 & \mci{35.3}{2.2} \\
\quad Bayes-UCB
  & 0.88 & 0.02 & \mci{43.8}{1.4}
  & 0.85 & 0.01 & \mci{33.8}{2.1} \\
\bottomrule
\end{tabular}
\captionof{table}{\textbf{Additional operational metrics.} SQL succ.\ is the fraction of SQL-requiring steps whose final query executes and returns rows; Attrition is the fraction of candidate regions excluded during search (``---'' if unused) when subquestion generation returns empty for that region. Valid Findings counts findings with positive rubric score (budget 50). Bootstrap CIs (half-width, 10{,}000 paired resamples) are shown for Valid Findings. The lower SQL success with \method methods is largely due to the agent erroneously attempting SQL execution even when operating in passage-only regions; among regions containing tables, SQL success is 98--100\%.
}
\label{tab:operational-main}
\end{center}

\textbf{Policies differ in how aggressively they discard regions.}
Random selection drops 23\% of its candidate regions to attrition on HybridQA, while the LLM policy drops none and the Bayesian policies only 1--3\% (Table~\ref{tab:operational-main}).
Yet the LLM policy scores lowest among \method variants on HybridQA ($0.35$): it reliably avoids regions that cannot answer anything, but concentrates on a small set of regions it judges safe and forfeits the coverage that makes even random selection strong.

\subsection{LLM Priors}

\begin{center}
    \small
    \setlength{\tabcolsep}{4pt}
    \begin{tabular}{llcrrrrr}
        \toprule
        \textbf{Policy} &
        \textbf{Priors} &
        \textit{Gr.} &
        \textit{Re.} &
        \textit{Di.} &
        \textit{Ut.} &
        \textbf{Score} \\
        \midrule
        \multirow{2}{*}{$\epsilon$-greedy}
            & No  & 0.893 & 0.686 & 0.663 & 0.618 & 0.308 \\
            & Yes & 0.899 & 0.768 & 0.806 & 0.697 & \textbf{0.437} \\
        \multirow{2}{*}{UCB}
            & No  & ---   & ---   & ---   & ---   & --- \\
            & Yes & 0.917 & 0.770 & 0.848 & 0.702 & \textbf{0.464} \\
        \bottomrule
    \end{tabular}
    \captionof{table}{\textbf{Effect of LLM priors (HybridQA).}
    Enabling LLM priors improves Baikal's $\epsilon$-greedy report score by $+0.129$.
    Bayes-UCB is the strongest policy evaluated.
    We do not report UCB without LLM priors because the research budget is much smaller than the number of selectable regions, so UCB would default to an unvisited (randomly ordered) region in each step and effectively behave like random exploration.}
    \label{tab:llm-priors-ablation}
\end{center}

\textbf{LLM priors are what make Bayesian search work.}
Without priors, Bayes $\epsilon$-greedy falls from $0.437$ to $0.308$ on HybridQA, a loss of $0.129$ that erases its entire advantage over the baselines (Table~\ref{tab:llm-priors-ablation}).
The budget is the reason: with $B=50$ steps against $|\mathcal{C}(q)| \approx 43$ candidate regions, a policy observes each region roughly once and cannot learn region values from rewards alone.

\begin{center}
\includegraphics[width=\textwidth]{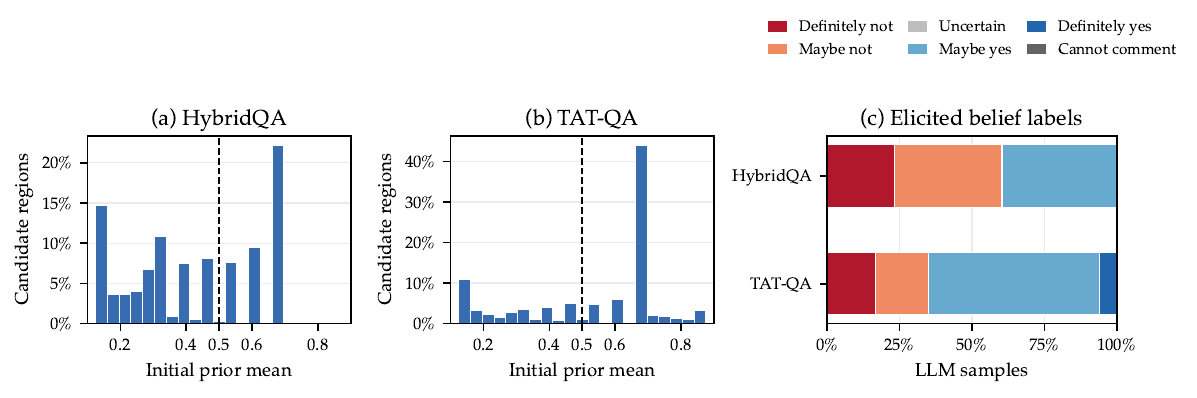}
\captionof{figure}{\textbf{Initial LLM priors over retrieved candidate regions.}
For each query, \method first retrieves the top-$k$ tables and passages by embedding similarity, retains each region overlapping this retrieval set, and elicits five categorical LLM judgments of whether that region could yield a grounded finding. These pseudo-Bernoulli samples are mapped to Beta priors for Bayes-UCB and Bayes $\epsilon$-greedy; panels (a--b) show the distribution of their initial means across all retained candidate regions (644 regions from 15 HybridQA queries; 642 from 15 TAT-QA queries). The dashed line denotes the uninformed prior mean of $0.5$. Panel (c) reports the underlying LLM belief labels. The heterogeneous distributions show that the LLM supplies discriminative, query-conditioned prior beliefs before any region is investigated.}
\label{fig:initial-priors}
\end{center}

Figure~\ref{fig:initial-priors} shows the priors carry the missing signal---elicited means are bimodal, concentrated near $0.15$ and $0.70$ with little mass at the uninformed value of $0.5$---so regions are separated before any of them is investigated.

\subsection{Retrieval Reachability}

\begin{center}
\small
\begin{tabular}{lcc}
\toprule
\textbf{Setting} & \textbf{Table Reach.} & \textbf{Passage Reach.} \\
\midrule
\multicolumn{3}{l}{\textit{Retrieval top-$k$ (using Qwen3-Embedding-0.6B)}} \\
\quad $k=25$  & 49.1 & 18.9 \\
\quad $k=50$  & 52.8 & 25.4 \\
\quad $k=100$ & 54.4 & 32.4 \\
\quad $k=200$ & 59.3 & 40.2 \\
\quad $k=400$ & 60.5 & 45.7 \\
\midrule[0.01pt]
\multicolumn{3}{l}{\textit{Embedding model (at $k=100$)}} \\
\quad Qwen3-Embedding-0.6B & 54.4 & 32.4 \\
\quad Qwen3-Embedding-4B   & 55.6 & 33.1 \\
\quad Qwen3-Embedding-8B   & 58.1 & 35.1 \\
\bottomrule
\end{tabular}
\captionof{table}{\textbf{Ground-truth reachability.} On TAT-QA, percentage of ground-truth tables and passages reachable through the retrieved regions induced by top-$k$ table/passage membership. Increasing retrieval depth ($k$) consistently improves coverage, while larger embedding models provide smaller additional gains at the default $k=100$. We choose defaults of $k=100$ and Qwen3-Embedding-0.6B based on computational convenience for this study.}
\label{tab:tatqa-retrieval-analysis}
\end{center}

\textbf{Retrieval depth matters more than embedding capacity.}
Two knobs determine which evidence is reachable at all---which gold items fall inside a region activated by the top-$k$ retrieval.
Raising $k$ from 25 to 400 on TAT-QA lifts table reachability from 49.1\% to 60.5\% and passage reachability from 18.9\% to 45.7\%, whereas scaling the encoder from 0.6B to 8B parameters at fixed $k=100$ adds only 3.7 and 2.7 points (Table~\ref{tab:tatqa-retrieval-analysis}).
Reachability also upper-bounds what any policy can find: at our default only about a third of gold passages are reachable, so a substantial share of \method's remaining error is inherited from the frontend rather than from search.

\subsection{LLM Scaling}

\begin{center}
    \includegraphics[width=0.72\textwidth]{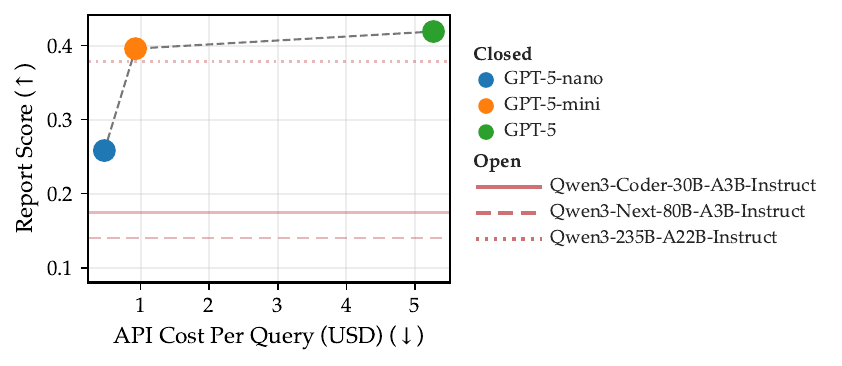}
    \captionof{figure}{\textbf{Model scaling improves quality with diminishing returns.}
    Average report quality versus API cost per query on the same three randomly sampled HybridQA queries.
    Among closed models, GPT-5 Mini captures most of the gain over GPT-5 Nano, while GPT-5 adds only a modest further improvement at substantially higher API cost.
    We also show scores for Qwen3 models as an estimate of open-weight performance (no additional API cost). Gemma-27B failed to produce any valid output, indicating that models need sufficient instruction-following and coding capability to work with our system.}
    \label{fig:model-scaling-cost-quality}
\end{center}

\textbf{LLM scale helps with steep diminishing returns, and \method runs on open weights.}
Across three HybridQA queries, moving from GPT-5-nano to GPT-5-mini lifts the report score from $0.26$ to $0.40$, while GPT-5 adds only about $0.02$ more at roughly six times the cost per query (Figure~\ref{fig:model-scaling-cost-quality}).
The largest open-weight model we test, Qwen3-235B-A22B-Instruct, comes within $0.02$ of GPT-5-mini at no API cost, whereas the 30B and 80B variants fall to $0.14$--$0.17$ and Gemma-27B produces no valid output at all.
\method therefore does not depend on a frontier model, but it does require an LLM with enough instruction-following and coding ability to drive the region-scoped agent.

\subsection{Rubric Agreement under Cohen's $\kappa$}

\begin{center}
\small
\setlength{\tabcolsep}{4.5pt}
\begin{tabular}{lcccc}
\toprule
\textbf{Dimension} & \textbf{Stat.} &
\textbf{Agreement ($\in [-1,1]$)} &
\textbf{Exact} & \textbf{Within-1} \\
\midrule
Groundedness & $\kappa$ & $0.44\ [0.26, 0.63]$ & 76.2\% & -- \\
Relevance & $\kappa_w$ & $0.13\ [-0.02, 0.28]$ & 41.2\% & 85.0\% \\
Distinctness & $\kappa_w$ & $0.61\ [0.32, 0.77]$ & 58.8\% & 92.5\% \\
Utility & $\kappa_w$ & $0.29\ [0.13, 0.43]$ & 18.8\% & 61.3\% \\
\bottomrule
\end{tabular}
\captionof{table}{\textbf{Cross-LLM rubric agreement under Cohen's $\kappa$ / $\kappa_w$.}
Same comparison as the main-text AC$_1$/AC$_2$ table: GPT-5-mini run-time judge vs.\ independent Claude Opus~4.6 on 80 score-stratified findings.
We report unweighted Cohen's $\kappa$ for groundedness and quadratic-weighted $\kappa_w$ for ordinal dimensions (95\% query-level bootstrap CIs).
Exact and Within-1 rates match the main-text table.
Because $\kappa_w$ is sensitive to skewed category prevalences, we use Gwet's $\mathrm{AC}_1$/$\mathrm{AC}_2$ as the primary agreement measures in the main text and include these estimates for transparency.}
\label{tab:llm-rubric-validation-kappa}
\end{center}

\clearpage

\section{Qualitative Case Study}
\label{app:qualitative-case-study}

\begin{lstlisting}[
    style=casequery,
    title={\bfseries Research Query}
]
How are @@population size, density, and recent demographic trends@@ associated with the @@quantity, types, and ages@@ (establishment/discovery dates) of cultural-heritage assets - including @@patterns of founding/sponsorship and other notable local attributes@@ - across Japanese municipalities designated as Cities of East Asia, core, or special cities and @@comparable non-Japanese localities@@ (U.S. cities with large Japanese-American populations, U.S. Marine Corps installations, cities in Balochistan, and Indian towns), and what @@regional patterns, temporal relationships, or plausible causal mechanisms@@ explain observed variations when accounting for @@designation timing and other contextual factors@@?
\end{lstlisting}

\begin{lstlisting}[
    style=casestudy,
    title={\bfseries \method{} (Bayes-UCB)}
]
Across the Japanese municipalities represented in the supplied tables, population size and density appear (*@\hlcase{associated with but do not fully determine}@*) the number, type, or age of cultural-heritage assets: mid-sized cities such as Miyakonojo (pop. 161,137; density 247/km2) (*@\hlcase{host multiple historic sites}@*) (Miyakonojo: 5 recorded sites across prefectural tables) while smaller municipalities (e.g., Hanamaki, pop. 94,127; density 104/km2) show lower household counts and (*@\hlcase{fewer recorded site entries}@*) in the sampled passages, illustrating variation at the municipal scale (population and density: P20750, P161887; site counts: T6179, T7301; household figures: P20750, P161887). Aggregate patterns show that (*@\hlcase{designation timing and administrative status correlate with population}@*): special-city designations in the 2000s encompass cities with average 2012 populations around 239k (T6194), and core-city listings (*@\hlcase{concentrate in historically larger regional centers}@*) (T6170), indicating that (*@\hlcase{statutory thresholds and administrative reclassification}@*) are important determinants of which municipalities concentrate large public stewardship responsibilities for heritage (T6194, T6170).

The types and ages of assets reflect (*@\hlcase{historical geography and founding/sponsorship patterns}@*) rather than contemporary demography alone: major concentrations of National Treasures and Heian/Kamakura-period structures are (*@\hlcase{clustered in Kyoto, Uji, Otsu, and other historical centers}@*) (National Treasures and writing/residence breakdowns; Heian structures by municipality: T3298/T6052/T7896-T7899/T9368; T10515/T6048), while the Hidden Christian sites are (*@\hlcase{concentrated in Nagasaki}@*) and are mostly 19th-20th-century in completion date (T6043, T8994; completion-date distribution: T6043/T8994). (*@\hlcase{Institutional and family sponsorship patterns}@*) explain many local stewardship arrangements: archives and libraries founded by aristocratic or ruling families (Yamauchi, Konoe, Hojo) and university, prefectural, or foundation affiliations account for provenance and collection ages (e.g., Tosa Yamauchi, Yomei Bunko, Kanazawa Bunko; Tenri Central Library; Kanagawa Prefectural Kanazawa-Bunko Museum) and recurring local patterns of family/industrial ownership appear in Okinawa structures and regional painting collections (founder/affiliation evidence: P90339, P215323, P168223, P132825; sponsorship patterns and ownership counts: T9113, T7573, T9365). Plausible causal mechanisms therefore combine (*@\hlcase{historical centrality, institutional capacity, and administrative thresholds}@*) rather than simple population density alone; data limits - e.g., missing or header-only location fields in some National-Treasure tables that prevent per-location age summaries and the lack of comparable non-Japanese site tables in the provided evidence - (*@\hlcase{constrain cross-national inference}@*) (limitations noted: inability to compute per-location age summaries: T3298/T7896 etc.; gaps for U.S./Marine/Balochistan/Indian comparisons).
\end{lstlisting}

\begin{lstlisting}[
    style=casestudy,
    title={\bfseries OpenCode Agent (+ Clustering)}
]
Across the sampled lake, (*@\hlcase{very old and high-value cultural assets cluster strongly in historical capitals}@*) and early-capital localities rather than in municipalities whose "core" or "special" administrative status is recent. (*@\hlcase{Kyoto and Nara prefectures dominate}@*) Heian- and Nara-period Important Cultural Property lists (Kyoto Heian entries, Nara-prefecture Nara entries) and are explicitly tied to imperial capital status and preserved prewar fabric [T10515; T6049; T5202; P213852; P3099]. (*@\hlcase{Small, low-population places can have extremely high densities}@*) of designated sites (Asuka village: approximately 20 Asuka-Fujiwara historic rows despite pop. approximately 5,681) consistent with archaeological/early-capital concentration, while (*@\hlcase{many recently designated cities do not intersect the ancient-asset tables}@*) in this sample - suggesting designation timing alone does not predict concentration of very ancient assets [T8750; P161979; T6194; T6170; T10515; T6049].

Quantitatively, passage-backed populations for six municipalities show (*@\hlcase{very different assets-per-capita}@*): Asuka (20 assets, pop. 5,681; 35.2 assets/10k), Hiraizumi (4 assets, pop. 7,440; 5.4/10k), Ikaruga (6, pop. 27,341; 2.2/10k), Sakata (8, pop. 101,521; 0.79/10k), Yamagata (6, pop. 249,743; 0.24/10k), Kyoto (6, pop. 1,474,473; 0.041/10k) [T8750; P161979; T10515; P144950; P28793; T7573; P221557; P56692; T6169; Finding 23]. (*@\hlcase{In this small n=6 sample}@*) there is a (*@\hlcase{weak negative association}@*) between population size and raw asset counts (Pearson r approximately -0.24; Spearman rho approximately -0.21), but the sample is limited, assets are raw unioned rows, and many municipal population fields in the city tables were non-numeric/parse-problematic - (*@\hlcase{constraints that limit inference}@*) (population parsing issue documented for T6169/T6170/T6194) [Finding 24; Finding 21]. Additional contextual patterns in the lake: (*@\hlcase{movable and painting ownership is mixed}@*) (private owners, museums, temples) [T7573]; Living National Treasure performing-arts designations (*@\hlcase{cluster in recent decades}@*) (1990s-2010s) [T10578]; and the non-Japanese comparison set shows (*@\hlcase{different temporal and founding patterns}@*) (early-20th-century ethnic districts such as Little Tokyo/Japantown and geographically dispersed Marine installations) rather than the ancient-capital signal seen in Japan [T6191; P118225; P185105; T9882].
\end{lstlisting}

\begin{lstlisting}[
    style=casestudy,
    title={\bfseries DeepSearcher}
]
Summary conclusion
- Broadly, population size, density, and recent demographic trends are (*@\hlcase{strongly correlated with the number, variety, and age-profile}@*) of cultural-heritage assets, but the relationship is mediated by historical role (capital, trade hub, military installation), legal/administrative status (designation as core/special city), disaster history, and local economic capacity. (*@\hlcase{Larger, denser, and growing municipalities tend to host more assets}@*); historically important but smaller cities concentrate very old and high-value heritage; purpose-built installations and rapidly rebuilt trade centres have predominantly modern heritage types.

Observed patterns and examples
- (*@\hlcase{Ancient capitals concentrate ancient, high-value heritage}@*): Nara illustrates how historical centrality produces a high density of ancient monuments. The Historic Monuments of Ancient Nara grouping includes eight sites (e.g., Todai-ji) and collectively contains numerous nationally designated assets (documents note 26 buildings as National Treasures and 53 Important Cultural Properties within the UNESCO properties). Todai-ji itself dates to the 8th century (founded 738 CE), showing how very early foundation dates produce concentrations of very old heritage irrespective of modern population size.
- (*@\hlcase{Mid-sized administrative cities show mixed portfolios}@*): Matsumoto (special city, slated to be a core city in 2020) had approximately 239,000 people (2019) and typifies municipalities that combine historical sites with modern civic infrastructure. (*@\hlcase{Administrative upgrades can increase heritage identification and protection}@*) through additional autonomy and resources.
- (*@\hlcase{Modern military installations generate recent heritage}@*): U.S. Marine Corps Base Camp Pendleton and its airfield date from World War II (commissioned/established 1942). Their heritage is typified by 20th-century military architecture, commemorative names, and operational landscape features rather than ancient religious or vernacular buildings.
- (*@\hlcase{Disaster histories reshape surviving asset stocks}@*): Quetta is a regional trade and communication centre; its major 1935 earthquake destruction and subsequent rebuilding means pre-1935 material heritage was lost or altered, producing a modern built environment with few surviving older monuments.
- (*@\hlcase{Diaspora concentrations influence cultural assets}@*): U.S. cities with large Japanese-American populations (not detailed in the supplied documents) typically host ethnic cultural institutions, festivals, and historic Japantown districts; such assets' presence and vitality are linked to historic migration waves, city size, and demographic persistence.

Regional, temporal and causal mechanisms
- (*@\hlcase{Historical centrality and founding date imply older heritage}@*). Cities that were political or religious centres early in history accumulate sacred complexes, palaces, and long-dated monuments.
- (*@\hlcase{Population size and density imply greater asset diversity}@*). Larger cities generate cultural infrastructure and possess resources to identify, maintain, and list heritage. However, sheer size does not guarantee ancient assets.
- (*@\hlcase{Administrative designation affects detection and preservation}@*). Core/special-city status can accompany expanded responsibilities and budgets that increase heritage inventories, protection, and tourism promotion.
- (*@\hlcase{Demographic decline can reduce maintenance capacity}@*), while growth can fund conservation but increase development pressure.
- (*@\hlcase{Disasters, conflict, and rebuilding shift asset age profiles}@*).
- (*@\hlcase{Functional role determines asset typology}@*): military installations produce memorials, barracks, and airfields; trade hubs produce mercantile buildings and markets; religious capitals have temples, shrines, and palace complexes.

Implications and research priorities
- Comparative inventories must control for founding age, historical role, disaster history, and administrative designation timing when relating population metrics to heritage counts.
- Useful future analyses include multivariate models, asset age-profile analyses, and comparisons between similarly sized cities with different histories.
- Preservation strategies should be tailored to ancient concentrations, growing urban areas, and modern military or post-disaster landscapes.

Limitations
- (*@\hlcase{The synthesis relies on illustrative examples and high-level patterns}@*). Full quantitative confirmation requires comprehensive, standardized inventories - including designation dates, establishment/founding years, and current demographic time series - across all target municipalities.
\end{lstlisting}

\begin{center}
\small
\begin{tabular}{lccc}
    \toprule
    \textbf{Metric} &
    \textbf{\method{} (Bayes-UCB)} &
    \textbf{OpenCode + Clustering} &
    \textbf{DeepSearcher} \\
    \midrule
    Report score      & \textbf{0.566} & 0.267 & 0.110 \\
    Valid findings    & \textbf{47}    & 23    & 6     \\
    Grounded          & \textbf{1.000} & 0.520 & 0.120 \\
    Relevance         & 0.830          & 0.425 & \textbf{0.965} \\
    Distinctness      & \textbf{0.805} & 0.375 & 0.695 \\
    Report usefulness & \textbf{0.795} & 0.400 & 0.130 \\
    \bottomrule
\end{tabular}
\end{center}

\captionof{figure}{%
\textbf{Qualitative case study.}
For a multi-faceted HybridQA query, \method{} integrates grounded
evidence about municipal demographics, designation timing, historical
geography, and institutional sponsorship. OpenCode + Clustering
identifies relevant historical patterns but supports part of its
synthesis using only six municipalities. DeepSearcher provides a broad
and highly relevant narrative, but many of its causal claims are not
grounded in retrieved evidence, reflected by its low groundedness and
report-usefulness scores. Highlighting marks the principal findings and
qualifications in each response.
}
\label{fig:qualitative-case-study}

\clearpage
\section{LLM Prompts}
\label{app:llm-prompts}

This appendix lists the LLM prompts used by \method{}, the finding-quality judge, and the OpenCode baselines.
Angle-bracket placeholders (e.g., \texttt{<user\_query>}) are filled at runtime.
Lines marked \texttt{[optional]} are emitted only under the stated condition.
Unless a listing gives a system message, calls use the default \texttt{You are a skilled data analyst.}

\subsection{Cluster Prior Elicitation}
\label{app:prompt-cluster-prior}

Used to sample categorical beliefs over each candidate region before Bayesian search.
The quality label matches the configured bandit reward:
\texttt{avg\_finding\_score}, \texttt{avg\_grounded\_relevance}, \texttt{avg\_grounded\_distinctness}, or \texttt{avg\_grounded\_usefulness}.

\begin{lstlisting}[style=llmprompt, title={\bfseries System}]
You are a creative and insightful research scientist.
\end{lstlisting}

\begin{lstlisting}[style=llmprompt, title={\bfseries User}]
Assess whether the given cluster of tables and passages could potentially be useful to answer the following research question.

RESEARCH QUESTION:
<user_query>

CLUSTER OF TABLES AND PASSAGES (id=<cluster_id>):
<cluster_context>

Could this cluster be helpful to answer the research question, primarily when assessing <quality_label>?

Return ONLY a valid JSON:
{"belief": "<one of: definitely yes, maybe yes, uncertain, maybe not, definitely not, cannot comment>"}
\end{lstlisting}

\subsection{Region Selection (LLM Policy)}
\label{app:prompt-region-selection}

\begin{lstlisting}[style=llmprompt, title={\bfseries User}]
You are helping build a deep-research report. Choose ONE candidate cluster to explore next.

RESEARCH QUESTION:
<user_query>

CANDIDATE CLUSTERS:
1. cluster_id=<id>
   visits=<n>, <reward_label>=<avg or n/a (never visited)>
<cluster_context>
...

Use your best judgment to select the next cluster to explore that may help discover a finding that improves on <reward_label>.

Return ONLY valid JSON:
{"selected_index": <1-based index>}
\end{lstlisting}

\subsection{Subquestion Generation}
\label{app:prompt-subquestion-generation}

The source note and answerability rule adapt to region contents:
\begin{itemize}\setlength{\itemsep}{2pt}\setlength{\parskip}{0pt}
\item \textbf{Tables + passages:} source = ``available tables (via SQL) and passages (text summaries)''; answerable using tables, passages, or both.
\item \textbf{Passages only:} source = ``available passages (text summaries)''; answerable using only the available passages.
\item \textbf{Tables only:} source = ``available tables using data retrieved from a simple or advanced SQL query''; answerable using only the available tables (may span multiple tables).
\end{itemize}
The ``already tried'' block appears only on revisits.
The empty-list rule appears only when empty candidate sets are allowed.

\begin{lstlisting}[style=llmprompt, title={\bfseries User (tables + passages variant)}]
Your task is to decompose the provided research question into smaller analytical sub-questions that can each be answered using the available tables (via SQL) and passages (text summaries). Make sure the sub-questions are phrased to be standalone -- the wording should contain all the relevant information needed to answer it, not being dependent on other sub-questions or the original research question or the list of available table/passage names.

RESEARCH QUESTION:
<user_query>

AVAILABLE CONTEXT:
<cluster_context>

[optional: only if the region was visited before]
ALREADY TRIED SUB-QUESTIONS FOR THIS CLUSTER:
- <previous_subquestion>
...

Do not repeat or rephrase any of the above (same intent counts as a duplicate). Generate only new, distinct sub-questions.

Return ONLY valid JSON:
{"subquestions": ["...", "...", ...]}

Requirements:
- Provide up to <K> diverse sub-questions relevant to the research question and the available context.
[optional: only when empty lists are allowed]
- If none of the available tables or passages are relevant to the research question, return an empty list.
- Each sub-question must be one sentence.
- Each sub-question must be answerable using the available tables, passages, or both. Some may require SQL over tables; others may be answerable from passages alone.
- Cover different analysis angles (counts, comparisons, trends, distributions).
- Do not repeat the full research question verbatim.
\end{lstlisting}

\subsection{Subquestion Deduplication}
\label{app:prompt-subquestion-dedup}

\begin{lstlisting}[style=llmprompt, title={\bfseries User}]
Your task is to remove questions from the "Candidate questions" list that are duplicate in meaning to one of the questions in the "Already answered" list.

Already answered:
- <answered_subquestion>
...

Candidate questions:
1. <candidate>
...

Return ONLY valid JSON:
{"remove_indices": [1, 3]}

List only the indices of candidate questions that need to be removed. Make sure to look for semantic equivalence to an already-answered question (same intent, not just similar wording). Return an empty list if none should be removed.
\end{lstlisting}

\subsection{Subquestion Selection (LSS)}
\label{app:prompt-subquestion-selection}

When no prior subquestions exist, the ``already asked'' block is the literal token \texttt{(none)}.

\begin{lstlisting}[style=llmprompt, title={\bfseries User}]
You are helping build a deep-research report that answers a research question.
Choose ONE candidate sub-question to investigate next.

RESEARCH QUESTION:
<user_query>

SELECTED CLUSTER CONTEXT:
<cluster_context>

SUB-QUESTIONS ALREADY ASKED:
(none)
  or
- <previous_subquestion>
...

CANDIDATE SUB-QUESTIONS:
1. <candidate>
...

Pick the sub-question most likely to produce a high-quality report finding:

1) Grounded -- answerable from the selected cluster's tables/passages (not speculative).
2) Relevance -- directly helps answer the research question.
3) Distinctness -- adds new information beyond sub-questions already asked.
4) Report usefulness -- worth including in a deep-research report (not redundant noise).

Return ONLY valid JSON:
{"selected_index": <1-based index>}
\end{lstlisting}

\subsection{SQL Need Decision}
\label{app:prompt-sql-decide}

Used by the lighter SQL-generation loop (ablation without the research agent).

\begin{lstlisting}[style=llmprompt, title={\bfseries User}]
You are deciding how to answer a question using available tables and passages.

Question:
<sub_question>

Available tables (SQL schema):
<schema_string>

Available passages:
<formatted_passages>

Return ONLY valid JSON:
{"needs_sql": true}

Set needs_sql to true only if answering the question requires querying the tables via SQL.
Set needs_sql to false if the passages alone are sufficient.
\end{lstlisting}

\subsection{SQL Generation}
\label{app:prompt-sql-generate}

\begin{lstlisting}[style=llmprompt, title={\bfseries System}]
You are a SQLite expert.
\end{lstlisting}

\begin{lstlisting}[style=llmprompt, title={\bfseries User}]
You are given a question and a schema. Your task is to write ONE SELECT query that answers the provided question using ONLY the tables and columns from the provided schema.

Schema:
<schema_string>

Example rows:
<samples_string>

Question: <sub_question>

Rules:
- Use only tables/columns from the provided schema.
- Output ONLY the SQL statement (no markdown, no other text).
- You may use aggregates (COUNT, GROUP BY, MIN, MAX, SUM, AVG, etc.), for instance, when the question asks for patterns, distributions, counts, rankings, comparisons, or ranges.
- When appropriate, you may use advanced SQL features (joins, subqueries, window functions, etc.).
- Make sure you sort the results by the most relevant column(s) when appropriate so that the results are easy to understand.
- Never invent tables or columns that are not present in the provided schema.
\end{lstlisting}

\subsection{SQL Refinement}
\label{app:prompt-sql-refine}

\begin{lstlisting}[style=llmprompt, title={\bfseries System}]
You are a SQLite debugging assistant.
\end{lstlisting}

\begin{lstlisting}[style=llmprompt, title={\bfseries User}]
You previously wrote a SQL query for a question, but it failed or returned 0 rows. Your task is to refine the query to fix the issue.

Schema:
<schema_string>

Example rows:
<samples_string>

Question:
<sub_question>

Previous SQL:
<previous_sql>

Observed issue:
- error_message: <error_message or (none)>
- empty_result (0 rows): <true|false>

Recent attempts:
- attempt <i>: ok=<bool> row_count=<n> error=<msg>
...

Rules:
- Use only tables/columns from the provided schema.
- Output ONLY the SQL statement (no markdown, no other text).
- You may use aggregates (COUNT, GROUP BY, MIN, MAX, SUM, AVG, etc.), for instance, when the question asks for patterns, distributions, counts, rankings, comparisons, or ranges.
- When appropriate, you may use advanced SQL features (joins, subqueries, window functions, etc.).
- Make sure you sort the results by the most relevant column(s) when appropriate so that the results are easy to understand.
- Never invent tables or columns that are not present in the provided schema.
- If the previous SQL returned an empty result, simplify the query.
- Prefer single-table unless there is a clear join key shared by tables.
\end{lstlisting}

\subsection{Answer from Passages and SQL}
\label{app:prompt-answer-context}

The SQL-results block is appended only when both a query and an execution result are present; on failure it is \texttt{SQL failed: <error>} instead of the success layout below.
The ``Tables referenced'' line appears only when table IDs can be extracted from the SQL.
Passage/SQL citation bullets are included only for evidence types that are present.

\begin{lstlisting}[style=llmprompt, title={\bfseries User}]
Answer the question using ONLY the provided passages and SQL results (if any). Be concise (2-5 sentences). Mention key numbers. If evidence is insufficient, say so.

Question: <sub_question>

Passages:
<formatted_passages>

[optional: only when SQL was executed]
SQL results:
SQL Query:
<sql>

[optional: only when table IDs are extracted]
Tables referenced: <table_ids>

Results (<row_count> rows):
<rows_preview>

Citation rules:
- Ground every factual claim in the provided evidence only.
- Do not invent facts beyond the evidence.
[optional: if passages are present]
- When a claim comes from a passage, cite its ID in brackets (e.g., [P42]).
[optional: if SQL results are present]
- When a claim comes from SQL results, cite the table ID(s) that supplied the data (e.g., [T7280]).
\end{lstlisting}

\subsection{Answer from SQL Only}
\label{app:prompt-answer-sql}

The ``Tables referenced'' line is omitted when no table IDs are extracted.
The ``showing up to \texttt{<max\_rows>}'' clause appears only when a row cap is set.

\begin{lstlisting}[style=llmprompt, title={\bfseries User}]
Answer the question using ONLY the SQL results below. Be concise (2-5 sentences). Mention key numbers. If results are empty, say so.

Question: <sub_question>

SQL Query:
<sql>

[optional: only when table IDs are extracted]
Tables referenced: <table_ids>

Results (<row_count> rows[, showing up to <max_rows>]):
<rows_preview>

Citation rules:
- Ground every factual claim in the provided evidence only.
- Do not invent facts beyond the evidence.
- When a claim comes from SQL results, cite the table ID(s) that supplied the data (e.g., [T7280]).
\end{lstlisting}

\subsection{Passage Expansion}
\label{app:prompt-passage-expansion}

Two variants decide whether to \texttt{grep} the passage index for additional evidence after a region-scoped investigation.
On SQL failure, the execution block is \texttt{SQL failed: <error>} rather than the success layout.

\begin{lstlisting}[style=llmprompt, title={\bfseries User (after SQL execution)}]
You are helping expand a research cluster with additional passages.

A sub-question will be answered using SQL over tables in the cluster. The SQL returned rows.
Decide whether searching the full passage description index with text keywords would help
answer the sub-question better or support follow-up analysis.

CURRENT CLUSTER:
<cluster_context>

SUB-QUESTION:
<sub_question>

SQL EXECUTION:
SQL Query:
<sql>

[optional: only when table IDs are extracted]
Tables referenced: <table_ids>

Results (<row_count> rows):
<rows_preview>

Return ONLY valid JSON:
{
  "expand": true,
  "grep_keywords": ["keyword or short phrase", "another keyword"],
  "generate_new_subquestions": false
}

Rules:
- Set expand to true only if grep keywords derived from the SQL results could surface relevant passages not already in the cluster.
- grep_keywords: 1-5 short search strings (OR semantics -- any match counts).
  Use concrete terms from SQL result values (names, places, events), not generic words.
- Set generate_new_subquestions to true only if the newly found passages would likely
  enable distinct follow-up sub-questions for this cluster.
- If expansion is not useful, return {"expand": false, "grep_keywords": [], "generate_new_subquestions": false}.
\end{lstlisting}

\begin{lstlisting}[style=llmprompt, title={\bfseries User (passage-only, no SQL)}]
You are helping expand a research cluster with additional passages.

A sub-question will be answered using only the passages currently in the cluster (no SQL).
Decide whether searching the full passage description index with text keywords would help
answer the sub-question better or support follow-up analysis.

CURRENT CLUSTER:
<cluster_context_without_passages>

SUB-QUESTION:
<sub_question>

CURRENT PASSAGES (only evidence available):
<formatted_passages>

Return ONLY valid JSON:
{
  "expand": true,
  "grep_keywords": ["keyword or short phrase", "another keyword"],
  "generate_new_subquestions": false
}

Rules:
- Set expand to true only if grep keywords could surface relevant passages not already in the cluster that would help answer the sub-question.
- grep_keywords: 1-5 short search strings (OR semantics -- any match counts).
  Use concrete terms from the sub-question and gaps in current passage coverage (entities, places, events mentioned in the question but weakly covered).
- Set generate_new_subquestions to true only if the newly found passages would likely
  enable distinct follow-up sub-questions for this cluster.
- If expansion is not useful, return {"expand": false, "grep_keywords": [], "generate_new_subquestions": false}.
\end{lstlisting}

\subsection{Final Report Synthesis}
\label{app:prompt-final-report}

\begin{lstlisting}[style=llmprompt, title={\bfseries User}]
Your task is to answer the provided research question by synthesizing the provided grounded sub-analyses as evidence into a cohesive report.

Research Question:
<user_query>

Evidence:
### Finding 1
Sub-question: <sub_question>
SQL Query: <sql>
Answer: <answer>
...

Synthesize the evidence into a cohesive report (1-<max_paragraphs> paragraphs). Cite which table/analysis supports each claim. Do not invent facts beyond the evidence. Output ONLY the report. No other text.
\end{lstlisting}

\subsection{Finding-Quality Rubric Judge}
\label{app:prompt-finding-rubric}

Used at run time to score each finding and as the independent cross-LLM validation judge.
Ordinal dimension guides and the absence-only override are expanded below as they appear in the filled prompt.
When there are no prior findings, that block is the literal token \texttt{(none)}.

\begin{lstlisting}[style=llmprompt, title={\bfseries System}]
You are a skilled data analyst evaluating the quality of research findings on data lakes.
\end{lstlisting}

\begin{lstlisting}[style=llmprompt, title={\bfseries User}]
Given a research question and a candidate finding on a data lake of tables and passages, your task is to evaluate the quality of the finding based on a provided rubric.

Research question:
"<user_query>"

Current finding:
Sub-question: <sub_question>
Answer: <answer>

Evidence for the finding:
Tables cited (SQL + answer): <tables_used>
Passages cited in answer: <passages_cited>
Cluster passages:
<cluster_passages>

Cited passage text:
<cited_passage_text>

SQL executed (<row_count> rows returned):
<sql>

SQL result preview:
<rows_preview>

Previously observed findings:
(none)
  or
1. Sub-question: <prior_sub_question>
   Answer: <prior_answer>
...

Important -- absence-only findings:
An answer is "absence-only" if it does NOT provide the factual information requested by the
sub-question, and instead only states that the information is missing, unavailable, not found,
not stated, not specified, or not present in the cited sources/SQL results. Examples:
- "Not stated in the provided documents."
- "No relevant information found."
- "The documents do not contain the peak chart position."

If the answer is absence-only, apply these overrides regardless of whether the absence claim
is technically supported by the cited evidence:
- grounded: "no" -- the finding does not deliver substantively grounded information that answers the sub-question
- report_usefulness: "none" -- documenting a search failure or data gap is not a useful report finding
- relevance and distinctness: score normally

Only score grounded "yes" when the answer provides the requested fact (e.g. a label, date, chart position,
certification level) supported by SQL and/or passage evidence.

Evaluate the current finding on four rubric dimensions.

1) Grounded in table or passage evidence?
Select exactly one:
- "yes" (1.0): The answer provides the factual information requested by the sub-question, and that information is supported by the SQL results and/or cited passage text
- "no" (0.0): The answer is absence-only (see above), unsupported, contradicted by evidence, or not tied to available evidence

2) Relevance to the research question domain?
For relevance, select exactly one label:
- "none" (0.0): Unrelated or does not help answer the research question
- "minimal" (0.25): Barely related; mostly off-topic for the research goals
- "partial" (0.5): Tangentially related but misses main analytical goals
- "substantial" (0.75): Mostly relevant with minor gaps
- "full" (1.0): Directly addresses an important part of the research question

3) Adds new information beyond previously observed findings?
For distinctness, select exactly one label:
- "none" (0.0): Duplicate or near-verbatim rephrase of a prior finding
- "minimal" (0.25): Mostly repeats prior findings with trivial wording changes
- "partial" (0.5): Overlaps heavily with prior findings but adds a small new detail
- "substantial" (0.75): Mostly new angle or evidence with some overlap
- "full" (1.0): Clearly new insight not covered by prior findings

4) Useful to include in a deep-research report (independent of other findings)?
For report usefulness, select exactly one label:
- "none" (0.0): Not worth including; noise, redundant in a report, or absence-only (only states that information was not found)
- "minimal" (0.25): Marginally informative; unlikely to help the reader
- "partial" (0.5): Somewhat helpful but low priority for the report
- "substantial" (0.75): Useful -- addresses requested aspects or adds complementary insight
- "full" (1.0): Highly useful -- directly answers part of the query or adds valuable complementary insight

Respond ONLY with a valid JSON:
{
  "grounded_reasoning": "<one sentence>",
  "grounded": "yes" or "no",
  "relevance_reasoning": "<one sentence>",
  "relevance": "<one of: none, minimal, partial, substantial, full>",
  "distinctness_reasoning": "<one sentence>",
  "distinctness": "<one of: none, minimal, partial, substantial, full>",
  "report_usefulness_reasoning": "<one sentence>",
  "report_usefulness": "<one of: none, minimal, partial, substantial, full>"
}
\end{lstlisting}

\subsection{OpenCode Full-Lake Agent}
\label{app:prompt-opencode}

Baseline agent that investigates the entire lake under a finding budget.
Shell commands are shown schematically; the implementation prefixes them with the lake tool entrypoint under the query working directory.
When passages are disabled, retrieval/cluster blurbs refer to tables only (no passage metadata), and passage citation examples are omitted.

\begin{lstlisting}[style=llmprompt, title={\bfseries User}]
You are a research agent answering a question using only the provided data lake.

Research question:
<user_query>

Query id: <query_id>

Schema descriptions: <schema_descriptions_path>
SQLite database: <sqlite_db_path>
Working directory: <query_dir>

[optional: if initial retrieval artifact is provided]
Initial retrieval candidates: <initial_retrieval_path>
This JSON lists embedding-ranked table/passage ids and titles as starting points.
Full schema and passage metadata remain in the files above.
  (tables-only variant: "table ids and titles"; "Full schema metadata remains...")

[optional: if initial clusters artifact is provided]
Initial inference clusters: <initial_clusters_path>
Topic-grouped table/passage clusters overlapping the retrieval candidates above.
Each cluster has cluster_id, description, tables, and passages.
  (tables-only variant: "table clusters"; "description, and tables")

Budget: record at most <budget> findings. Each finding is one evidence-gathering step.
Per finding you may run up to <max_sql_attempts> SQL attempts before committing it.

Run SQL (table names may be unquoted or double-quoted, e.g. T809 or "T809"):
<lake_tool> sql "SELECT * FROM T809 LIMIT 10"

Record a finding (required after each step):
<lake_tool> commit --sub-question "..." --answer "..."
<lake_tool> commit --sub-question "..." --answer "..." --no-sql

Citation rules (required in every --answer):
- Ground every factual claim in lake evidence only.
- When a claim comes from SQL results, cite the table ID(s) in brackets (e.g., [T809]).
- SQL-backed commit example:
  <lake_tool> commit --sub-question "What is the stadium capacity?" --answer "The capacity is 50,000 [T809]."
[optional: if passages are enabled]
- When a claim comes from a passage, cite its ID in brackets (e.g., [P42]).
- Passage-only commit example:
  <lake_tool> commit --sub-question "What does the passage say about density?" --answer "Urban density is higher in the core [P42]." --no-sql

Finding quality rules (required):
- Do not commit findings that only report table row counts (e.g. SELECT COUNT(*) FROM T...).
- Do not spend findings on schema discovery (sqlite_master, listing tables, column introspection).
- Each finding must contain substantive analysis that advances the research question.
[optional: if initial retrieval artifact is provided]
- Start by looking at the initial retrieval candidates above before exploring the full lake.

Check budget status:
<lake_tool> status

[optional: if passages are enabled]
Passage descriptions: <passage_descriptions_path>
Load a passage: <lake_tool> passage P42

Use only lake evidence. Do not use outside knowledge. Do not invent facts beyond the evidence.
Do not modify the lake in any way.

Autonomous execution (required):
- You are running fully unattended; no human is available to answer questions.
- Never ask questions, present option menus, or end a turn waiting for user input.
- If multiple next steps are possible, choose the one most likely to gather evidence and execute it immediately.
- Keep running sql/commit cycles until `status` reports the budget is exhausted (<budget> findings).
\end{lstlisting}

\subsection{OpenCode Continuation and Resume}
\label{app:prompt-opencode-continuation}

Issued when the agent pauses for user input or a session is resumed mid-budget.

\begin{lstlisting}[style=llmprompt, title={\bfseries Continuation}]
No human is available. Proceed autonomously using your best judgment.

Your last output was:
<agent_continuation_context>

Do not ask further questions or present option menus. Choose the best course of action yourself and continue executing sql/commit cycles until `status` reports the budget is exhausted.
\end{lstlisting}

\begin{lstlisting}[style=llmprompt, title={\bfseries Resume}]
No human is available. Resume this research task autonomously.

Progress: <n_findings>/<budget> findings committed. Budget step is <step>. <remaining> finding(s) remain.

Findings already committed are recorded in lake_tool_state.json -- do not redo them.
Continue from the current step with sql/commit cycles until `status` reports the budget is exhausted.

Do not ask questions or present option menus. Execute the next evidence-gathering step immediately.
\end{lstlisting}

\subsection{OpenCode Subquestion Executor (Research Agent)}
\label{app:prompt-opencode-subquestion}

Region-scoped coding agent used by full \method{} configurations (RA).
The passage-index block is included when passages are enabled; the expansion block only when passage expansion is also enabled.

\begin{lstlisting}[style=llmprompt, title={\bfseries User}]
You are a research agent answering ONE sub-question for a deep-research report.

Research question:
<user_query>

Sub-question to answer (use this exact text in commit --sub-question):
<sub_question>

SELECTED CLUSTER (tables and passages in scope):
<cluster_context>

Cluster schema (SQLite contains ONLY these tables: <table_ids>):
<cluster_schema_path>
SQLite database: <sqlite_db_path>
Working directory: <query_dir>

[optional: if passages are enabled]
Passage descriptions index: <passage_descriptions_path>
Load a passage by id: <lake_tool> passage P42

[optional: if passage expansion is enabled]
Optional cluster expansion (before you commit):
- If cluster passages are insufficient, you MAY search for additional passages.
- When tables exist and you run SQL: inspect SQL results FIRST, then grep using
  concrete terms from returned values (names, places, events -- not generic words).
- When answering passage-only without SQL: grep from the sub-question and gaps
  in current passage coverage.
- Search (max <MAX_ADDED_PASSAGES> new passages per grep, OR keyword matching):
  <lake_tool> grep-passages --keywords "keyword" "another"
- Load promising hits with passage before citing them in your answer.
- Do not grep before reviewing SQL results when SQL was used.

You have budget for exactly ONE finding. Run up to <max_sql_attempts> SQL attempts, then commit.

Run SQL (table names may be unquoted or double-quoted):
<lake_tool> sql "SELECT * FROM T809 LIMIT 10"

Record your finding (required -- use the sub-question above verbatim):
<lake_tool> commit --sub-question "<sub_question>..." --answer "Answer with citations [T809] or [P42]."
<lake_tool> commit --sub-question "<sub_question>..." --answer "Passage-only answer [P42]." --no-sql

Citation rules (required in every --answer):
- Ground every factual claim in lake evidence only.
- Cite table IDs in brackets for SQL-backed claims (e.g., [T809]).
- Cite passage IDs in brackets for passage claims (e.g., [P42]).

Check status:
<lake_tool> status

Use only lake evidence. Do not use outside knowledge. Do not modify the lake.

Autonomous execution (required):
- You are running fully unattended; no human is available.
- Never ask questions or wait for user input.
- Commit exactly one finding for the sub-question above, then stop.
\end{lstlisting}

\fi

\end{document}